\documentclass[5p]{elsarticle}
\usepackage[utf8]{inputenc} 
\usepackage[T1]{fontenc}    
\usepackage{url}            
\usepackage{booktabs}       
\usepackage{amsfonts}       
\usepackage{nicefrac}       
\usepackage{microtype}      
\usepackage{graphicx} 
\usepackage{amsmath, amsthm, amssymb}
\usepackage{tabularx}
\usepackage{float}
\usepackage{adjustbox}

\DeclareMathOperator\arcsinh{arcsinh}
\DeclareMathOperator\asinh{asinh}

\newtheorem{theorem}{Theorem}
\graphicspath{ {images/} }
\usepackage{color}

\def\t{{\mathbf{t}}}
\def\x{{\mathbf{x}}}
\def\N{{\mathbb{N}}}
\def\u{{\mathbf{u}}}
\def\cX{{\mathcal{X}}}
\def\y{{\mathbf{y}}}
\def\cY{{\mathcal{Y}}}
\def\GP{{\mathcal{GP}}}
\def\O{{\mathcal{O}}}
\def\R{{\mathbb{R}}}
\def\cN{{\mathcal{N}}}

\def\sgn{{\text{sgn}}}

\journal{Journal of Neural Networks (author copy, accepted 22/6/2019)}
\begin{document}

\begin{frontmatter}

\title{Compositionally-Warped Gaussian Processes}

\author[DIM]{Gonzalo~Rios}
\ead{grios@dim.uchile.cl} 

\author[DIM,CMM]{Felipe~Tobar\corref{mycorrespondingauthor}}
\ead{ftobar@dim.uchile.cl}
\cortext[mycorrespondingauthor]{Corresponding author}

\address[DIM]{Department of Mathematical Engineering, Universidad de Chile. }
\address[CMM]{Center for Mathematical Modeling, Universidad de Chile. \\Beauchef 851, 8370456, Santiago, Chile.}

\address{Universidad de Chile}


\begin{abstract}
	The Gaussian process (GP) is a nonparametric prior distribution over functions indexed by time, space, or other high-dimensional index set. The GP is a flexible model yet its limitation is given by its very nature: it can only model Gaussian marginal distributions. To model non-Gaussian data, a GP can be warped by a nonlinear transformation (or \textit{warping}) as performed by warped GPs (WGPs) and more computationally-demanding alternatives such as Bayesian WGPs and deep GPs. However, the WGP requires a numerical approximation of the inverse warping for prediction, which increases the computational complexity in practice. To sidestep this issue, we construct a novel class of warpings consisting of compositions of multiple elementary functions, for which the inverse is known explicitly. We then propose the compositionally-warped GP (CWGP), a non-Gaussian generative model whose expressiveness follows from its deep compositional architecture, and its computational efficiency is guaranteed by the analytical inverse warping. Experimental validation using synthetic and real-world datasets confirms that the proposed CWGP is robust to the choice of warpings and provides more accurate point predictions, better trained models and shorter computation times than WGP.
\end{abstract}


%

\end{frontmatter}

 
\section{Introduction}
\label{sec:intro}

\subsection{The Gaussian distribution} 
\label{sub:suitability}

	
Two main reasons for the extensive use of the Gaussian distribution in science can be identified: one is conjectural and the other one is practical. The first reason obeys the simplifying assumptions in mathematical modelling, since observed data comprise multiple error-corrupted phenomena, an exact mathematical description of these real-world data-generating engines is challenging---if not impossible. Therefore, we partially model the data using first principles to then describe the remaining components as several sources of uncertainty added together, i.e., the \textit{noise}. Then, based on the central limit theorem \cite{araujo1980central}, we can define  this so-called noise in statistical terms by a Gaussian distribution. 

The second reason is the appealing mathematical properties of the Gaussian distribution, in particular, for Bayesian inference and learning \cite{rasmussen06}. Gaussian random variables (RVs) are closed under \textit{conditioning} and \textit{marginalisation}, i.e., all marginal and conditional distributions of a set of jointly-Gaussian RVs are Gaussian; this allows for tractable inference. Additionally, as Gaussian distributions are also closed under multiplication, they are conjugate priors for themselves, meaning that a Gaussian prior and likelihood result in a Gaussian posterior distribution. This closed-form posterior allows for (i) efficient gradient-based learning via optimisation, and (ii) exact Bayesian inference. 

There is a number of models that rely on the Gaussian distribution, even when the data are known to be non-Gaussian but Gaussianity is assumed to avoid the computational complexity related to more realistic models---see e.g. the use of the Kalman filter in the Apollo missions \cite[Sec. 1]{crisan09}. We now motivate the construction of a Gaussian process, i.e., a model for infinitely-many jointly-Gaussian random variables, using a neural network with an infinite number of neurons.

\subsection{From neural networks to Gaussian processes} 
 \label{ssub:NN2GP}
 
 \begin{figure}
	\centering
	\includegraphics[width=0.4\textwidth]{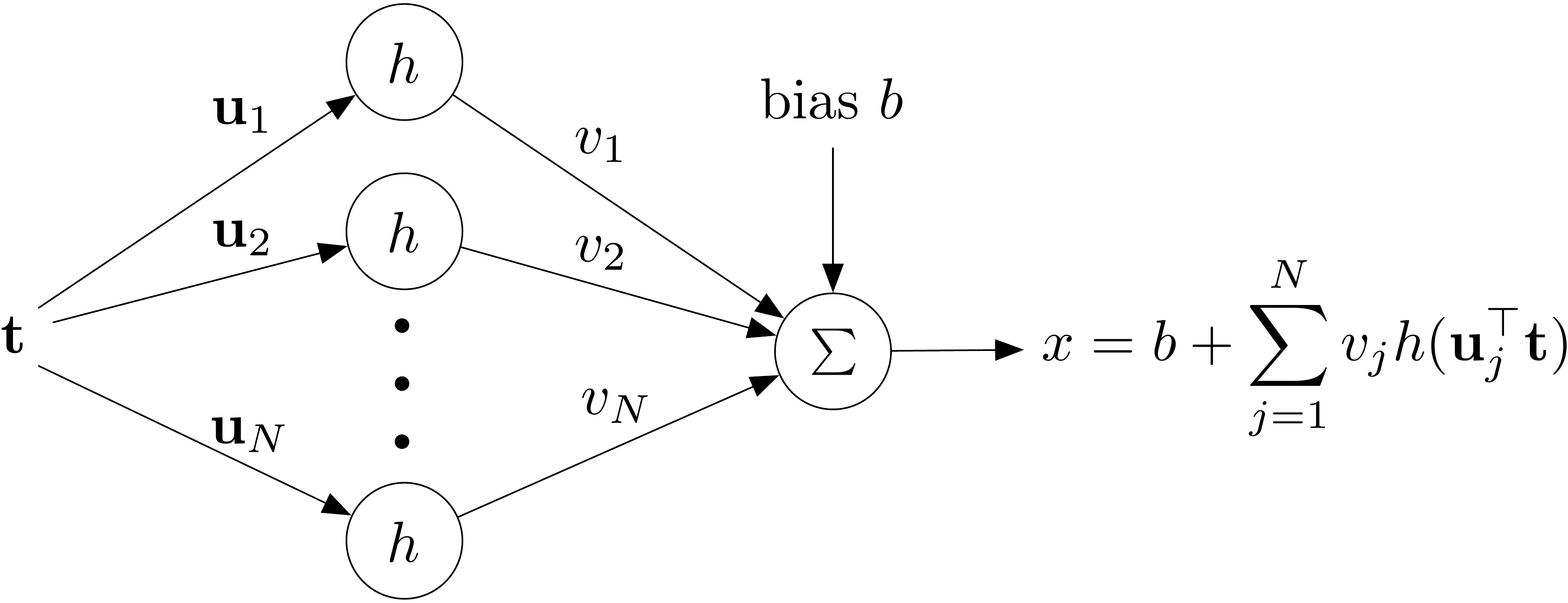}\\
	\caption{Single-layer feedforward neural network: $\t$ is the input, $x$ is the output, $h(\cdot)$ is the activation function, $b$ is the bias, $\u_{j=1:N}$ are the input weights, $v_{j=1:N}$ are the output weights. }
	\label{fig:NN}
\end{figure}

Among neural network practitioners, it is widely believed that the number of neurons should be determined based on the amount of available data. However, as pointed out by C.~Williams in \cite{NIPS1996_1197}, this makes little sense from a Bayesian standpoint, where the complexity of the model should be dictated by the complexity of the problem and not by the amount of available data. In this regard, R.~Neal demonstrated that the output of a single-layer neural network with random weights converges to Gaussian process when the number of neurons approaches infinity \cite{Neal:ARD}. 

Following \cite{rasmussen06,NIPS1996_1197,Neal:ARD}, let us consider a single-layer $N$-neuron neural network as shown in Fig.~\ref{fig:NN}. By modelling the bias and weights as independent random variables, the outputs $x_1,x_2,\ldots,x_N$ are also random for any choice of inputs $\t_1,\t_2,\ldots,\t_N$, with a distribution that is not necessarily tractable due to the nonlinear activation function $h(\cdot)$. Nevertheless, notice that the network in Fig.~\ref{fig:NN} is defined by a sum of i.i.d. terms, therefore, by virtue of the multidimensional central limit theorem (CLT \cite{araujo1980central}), taking the number of neurons $N\rightarrow\infty$ results in the outputs $x_1,x_2,\ldots,x_N$ being jointly Gaussian\footnote{The motivation for taking the number of neurons to infinity follows \cite{Hornik:1993}, which states that the network in  Fig.~\ref{fig:NN} is a universal approximator. Furthermore, the CLT can in fact be applied since the bounded activation function $h$ results in finite variance for the outputs $x_1,x_2,\ldots,x_N$. Notice that scaling the output weights variance $\propto 1/N$ is required for the CLT to hold.}. This construction can be further extended to the case of an infinite number of outputs via the Kolmogorov consistency theorem \cite{tao2011introduction}, thus yielding the \textit{Gaussian process} \cite{rasmussen06}.


\subsection{Shortcomings of Gaussian processes} 
\label{sub:shortcomings}

Despite the facts in favour of the Gaussian distribution presented in Section \ref{sub:suitability}, the assumption of joint Gaussianity is far from reality in a number of settings. In practice, one deals with observations that are non-symmetric, heavy-tailed, always positive, or bounded by a physical or economic restriction; all of these properties are contradictory with the Gaussian framework. In order to model non-Gaussian data while still making use of the advantages of Gaussian models, one can regard the data as a nonlinear transformation of jointly-Gaussian RVs. This is a standard approach in Statistics, where classic examples of transformations are the logarithmic and Box-Cox functions \cite{sakia1992box}. 

\begin{figure}
	\centering
	\includegraphics[width=0.4\textwidth]{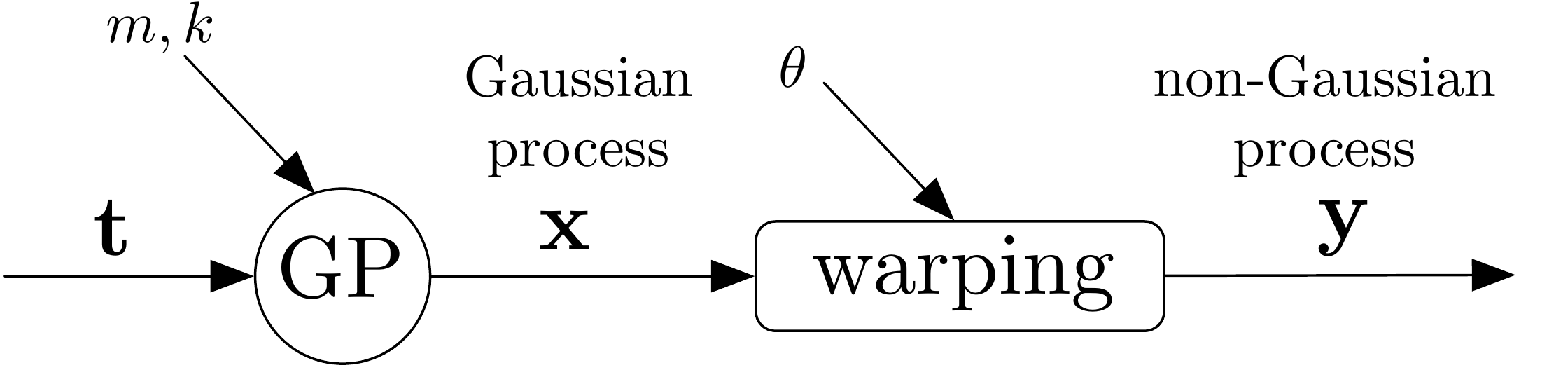}\\
	\caption{General structure of warped Gaussian processes where a GP is nonlinearly transformed to model non-Gaussian observations. The warping can be a sum of $\tanh(\cdot)$ functions (standard WGP), another GP (Bayesian WGP), a composition of GPs (deep GP), or a composition of invertible and differentiable warping as in the proposed model (CWGP). Computing predictions using this model requires us to evaluate of the inverse warping; which is known in closed form for the proposed CWGP.}
	\label{fig:wgp}
\end{figure}

To model non-Gaussian processes, a transformation referred to as \textit{warping} can be applied to a GP---see Fig.~\ref{fig:wgp}. This warping can be given by a sum of hyperbolic tangent functions as in warped GP (WGP \cite{warped04}), by another GP as in Bayesian WGP (BWGP \cite{bayesianwarped12}), or by a sequence of GPs as in deep GP (DGP \cite{deep_GP}). The main drawback of these approaches is the computational complexity arising from both training and prediction due to intractability of the model. In particular, WGP, which has the lowest computational complexity of the three, requires to numerically approximate the inverse of the warping function for prediction. Our aim is to construct a novel warping for Gaussian processes that inherits the expressiveness of deep structures but at the same time require minimal numerical approximations for prediction; this will be attained by constructing warpings with known closed-form inverse.

\subsection{Proposal and contributions} 
\label{sub:our_proposal_}

This paper presents a generative model for non-Gaussian processes, where a (latent) Gaussian process is \textit{passed through} an \textbf{invertible} and \textbf{differentiable} nonlinear warping. We achieve this by constructing warping functions through the composition of multiple (elementary) functions. By choosing elementary functions with derivatives and inverses known in closed form, both the inverse and derivative of the warping are known in closed form too due to the properties of the composition operation. The proposed model, termed compositionally-warped GP (CWGP), can then be interpreted as a neural network with an infinitely-wide first layer (the GP) fed into a deep structure comprising several single-neuron layers (the warping).  

Our contributions are: 
\begin{itemize}
	\item[i)] A reinterpretation of the construction of warped Gaussian processes from a measure theoretic perspective.
	\item[ii)] A principled procedure to transform GPs using a composition of elementary functions. This construction requires minimal numerical approximations with respect to existing methods due to the existence of closed-form formulae for prediction and learning.
	\item[iii)] A set of elementary functions with closed-form derivatives and inverses together with a recommendation on how to use them, regardless of the availability of the data's statistical properties.
	\item[iv)] A study of the appealing computational complexity of the proposed CWGP method both in theoretical and experimental terms.
	\item[v)] Illustrative examples using synthetic and real-world data that validate the proposed method against WGP in terms of replicability, computational efficiency, and predictive ability.
\end{itemize}

\subsection{Structure summary} 
\label{sub:structure}
The article is organised as follows. Sec.~\ref{sec:background} introduces the required theoretical background and related work. Sec.~\ref{sec:novelWGP} presents and analyses the compositionally-warped Gaussian process (CWGP), which is the main contribution of the article. Then, Sec.~\ref{sec:transformations} describes specific transformations for CWGP and highlights their properties, while Sec.~\ref{sec:choice_trans} advises on how to use them in different scenarios. Sec.~\ref{sec:experiments} validates the proposed CWGP against standard GPs and warped GPs in three real-world scenarios in terms of training accuracy, generalisation and computation time. Lastly, Sec.~\ref{sec:discussion} concludes the article and proposes future research directions based on our findings.


\section{Nonlinear transformations of Gaussian processes}
\label{sec:background}
\subsection{Notation} 

Recall that the (multivariate) distribution of a jointly-Gaussian random vector $\x \in \R^{n}$ with mean $\mu \in \R^{n}$ and covariance matrix $\Sigma \in \R^{n\times n}$ is given by
\begin{eqnarray*}
	\mathcal{N}\left(\mathbf{x}| \mathbf{\mu},\Sigma \right)&=&\frac{1}{\left( 2\pi \right) ^{
			\frac{n}{2}}\left\vert \Sigma \right\vert ^{\frac{1}{2}}}e^{-\frac{1}{2}
		\left( \mathbf{x}-\mathbf{\mu}\right) ^{\top }\Sigma ^{-1}\left( \mathbf{x}-\mathbf{\mu}
		\right) },
\end{eqnarray*}
where $\left\vert \Sigma \right\vert$ denotes the determinant of the matrix $\Sigma$. Due to its consistency under both marginalisation and permutation, we can extend the finite-dimensional multivariate Gaussian distribution to the infinite-dimensional case through Kolmogorov's consistency theorem \cite{tao2011introduction}. This construction is referred to as the Gaussian process (GP) \cite{rasmussen06}, a prior probability distribution over functions that defines nonlinear nonparametric regression models by assuming joint Gaussianity of the observed data.

We use the notation $x\left( t\right) \sim \mathcal{GP}\left( m(t),k\left( t,\bar{t}\right)\right)$ for a GP indexed by the variable $t\in \mathcal{T}$, $\{x(t)\}_{t\in \mathcal{T}}$, with mean function $m(\cdot)$ and covariance function $k(\cdot,\cdot)$, where the parameters of $m$ and $k$ are referred to as \emph{hyperparameters} of the GP. For a finite collection of points in the domain of the process $\t=[t_1, \ldots, t_N]^\top \in \mathcal{T}^N$, we denote the vector $x(\t) = [x(t_1), \ldots, x(t_N)]^\top\in\mathbb{R}^N$, which follows a multivariate Gaussian distribution with mean vector $m(\t)= [m(t_1), \ldots, m(t_N)]^\top$ and covariance matrix $[k(\t,\t)]_{ij} = k(t_i,t_j)$. As long as there is no ambiguity in the choice of points $\t$, we will denote $x(\t)$ as $\x$, $m(\t)$ as $\mu_{\x}$ and $k(\t,\t)$ as $\Sigma_{\x}$. For a second collection of input points $\t'$ the notation is analogue: the process evaluation is $\x'=x(\t')$, the mean is $\mu_{\x'}=m(\t')$ and the cross-covariance is $\Sigma_{\x\x'}=k(\t,\t')$.

\subsection{The change of variables theorem and existence of a non-Gaussian process}

 A standard approach to model non-Gaussian observations is to transform the data using, e.g., the logarithmic \cite{boxcox} or hyperbolic tangent \cite{johnsonsu} functions, so that the transformed data are (closer to being) normally distributed. This transformation results in a change of probability measure \cite{tao2011introduction}, where the distribution of the transformed variable is known explicitly given the transformation. However, this result and its theoretical implications in the construction of expressive non-Gaussian models are usually neglected. We will now formally present the change of probability measure resulting from transforming a random variable  via the following theorem and then study the Gaussian case.

\begin{theorem}
\label{thm:CoV}
	(Probability change of variables \cite{hogg1995introduction}) Let  $\x \in \cX \subseteq \R^{n}$ be a random vector with a probability density function given by $p_{\x}\left(\x\right)$, and let $\y\in \cY\subseteq \R^n$ be a random vector such that $\varphi \left(\y\right) = \mathbf{x}$, where the function  $\varphi :\cY\rightarrow \cX$ is bijective of class $\mathcal{C}^{1}$ and $\left\vert \nabla \varphi\left( \y\right) \right\vert >0$ $\forall \y\in \cY$. 

	Then, the probability density function $p_\y(\cdot)$ induced in $\mathcal{Y}$ is given by 
	\begin{equation*}
	p_{\y}\left( \y\right) =p_{\x}\left( \varphi \left( \y\right)\right) \left\vert \nabla \varphi \left( \y\right) \right\vert,
	\end{equation*}
	where $\nabla \varphi \left( \cdot\right)$ denotes the Jacobian of $\varphi \left( \cdot\right)$, and recall that $|\cdot|$ denotes the determinant operator.
\end{theorem}
We refer to $\x=[x_{1},...,x_{n}]^\top$ as the \emph{base} variables and to $\y=[y_{1},...,y_{n}]^\top$ as the \emph{transformed} variables. The change of variables theorem gives a principled methodology to express the probability density function (pdf) of the transformed variables in terms of (i) the pdf of the base variables and (ii) the applied transformation. 

As our aim is to use the change of variables theorem to construct non-Gaussian tractable models, let us consider a multivariate normal random vector $\x\in\R^n$ with mean $\mu_{\x}$ and covariance $\Sigma_{\x}$, denoted by $\x\sim\cN(\mu_{\x},\Sigma_{\x})$, and a coordinate-wise\footnote{To simplify the notation we refer to both the vector or scalar maps indistinctly as $\varphi$.} mapping from the transformed space to the base space given by 
\begin{equation*}
 \y \mapsto \x=\varphi(\y) = [\varphi(y_{1}),...,\varphi(y_{n})]^\top.	
\end{equation*}
 Notice that the Jacobian of $\varphi(\y)$ is diagonal and therefore its determinant factorises as 
\begin{equation*}
\left\vert \nabla \varphi \left( \y\right) \right\vert = \prod_{i=1}^{n}\frac{d\varphi\left( y_{i}\right) }{dy} > 0.
\end{equation*}
In this setting, the pdf of $\y = [y_1,\ldots,y_N]^\top$ can be obtained explicitly through Theorem \ref{thm:CoV} and takes the form
 \begin{align*}
 	p(\y) &=\prod\limits_{i=1}^{n}\frac{d\varphi\left( y_{i}\right) }{dy}\mathcal{N}\left( \varphi(\y) |\mu_{\x},\Sigma_{\x}\right),
 \end{align*}
where the function $\varphi$ is affine if and only if the distribution $p(\y)$ is Gaussian. In general, the distribution $p(\y)$ is not Gaussian, yet it is parametrised by the base mean $\mu_{\x}$, the base variance $\Sigma_{\x}$ and the transformation $\varphi$.

Theorem \ref{thm:CoV} can also be used to calculate conditional densities of transformed Gaussian random vectors: For two jointly-Gaussian vectors $\x,\x'$ with conditional density $p(\x\vert \x')=\cN\left( \mu_{\x|\x'} , \Sigma_{\x|\x'}\right)$, and a pair of vectors $\y,\y'$ such that $\x=\varphi(\y)$ and $\x'=\varphi(\y')$, the conditional density $p(\y\vert\y')$ is given by  
\begin{eqnarray*}
	p\left( \y|\y'\right) &=&\prod\limits_{i=1}^{n}\frac{d\varphi\left( y_{i}\right) }{dy} \cN\left( \varphi\left( \y\right) | \mu_{\x|\x'} , \Sigma_{\x|\x'}\right)\\
	\mu_{\x|\x'} &=& \mu_{\x}+\Sigma _{\x\x'}\Sigma _{\x'\x'}^{-1}\left( \varphi\left( \y'\right) -\mu_{\x'}\right)\\
	\Sigma_{\x|\x'} &=& \Sigma _{\x\x}-\Sigma _{\x\x'}\Sigma _{\x'\x'}^{-1}\Sigma _{\x'\x},
\end{eqnarray*}
where recall that  $\Sigma _{\x\x'}$ denotes the covariance between $\x$ and $\x'$, and $\mu_\x$ denotes the marginal mean of $\x$. 

Observe that the posterior density of the transformed element $p\left( \y|\y'\right)$ belongs to the same family as the unconditional density $p(\y)$. This property of closure under conditioning is inherited from the (base) Gaussian pdf and it is preserved by the coordinate-wise transformation $\varphi$. Furthermore, the non-Gaussian multivariate distribution $p(\y)$ is also closed under marginalisation and permutation, again since $\varphi$ is defined coordinate-wise. 

Therefore, we can construct a non-Gaussian process by \emph{transforming} (or \emph{warping}) a GP in the following manner: (i) choose a base GP $x$ and a coordinate-wise transformation $\varphi$, (ii) compute the finite-dimensional marginal densities of $y$ s.t. $x=\varphi(y)$ via the change of variable theorem, and (iii) apply the Kolmogorov consistency theorem \cite{tao2011introduction}. This construction guarantees the existence of such non-Gaussian process with known hyperparameters: the mean and covariance of the base GP and the transformation $\varphi$. 

\subsection{Prior art: Warped Gaussian processes}

Warped Gaussian processes (WGP) \cite{warped04} follow the rationale explained in the previous section. WGP considers a GP with zero mean and square-exponential (SE) covariance function, as well as a monotonic (and thus invertible) parametric coordinate-wise transformation. 

The transformation $\varphi: \R\rightarrow\R$ considered by WGP is given by
\begin{equation}
	\label{eq:wgp}
	\varphi \left( y\right)  =y+\sum_{j=1}^{d}a_{j}\tanh \left( b_{j}\left(
	y+c_{j}\right) \right),
\end{equation}
where $a_{j},b_{j} \geq 0,\ j=1,\ldots,d$. The mixture of the identity and hyperbolic tangent functions in eq.~\eqref{eq:wgp} acts as a parametric {warping} of the identity function, meaning that standard transformations such as a the logarithm are not allowed by WGP. Observe that since $\varphi \left( y\right)$ in eq.~\eqref{eq:wgp} is a sum of monotonic terms, its inverse does exists. However, as this inverse it not known explicitly, computing the predictive posterior WGP requires approximating $\varphi^{-1}$ using, e.g., the Newton-Raphson method (NRM) \cite{atkinson2008introduction}. This iterative procedure requires several evaluations of  $\varphi$ and $\frac{d\varphi}{dy}$, thus increasing computational complexity, in addition to being sensitive to the initial condition. In practice, the use of NRM is the computational bottleneck of WGP: the original model proposed in \cite{warped04} considered a na\"ive NRM approach that resulted in inference being one or two orders of magnitude more expensive than that of standard GPs. For computational efficiency, the implementation of \cite{warped04} considered a bisection search to find appropriate initial conditions for NRM. We emphasise that although the implementation of WGP can be made more efficient by using sophisticated numerical tools for approximating inverse functions, e.g., to train a surrogate model for the inverse using splines or neural networks, WGP always requires numerical approximations when performing predictions due to the lack of the explicit inverse of a sum of hyperbolic tangents. On the contrary, the model to be proposed in Sec.~\ref{sub:model_description} does not suffer from this drawback.


A non-parametric version of WGP is the Bayesian WGP \cite{bayesianwarped12}, denoted BWGP, which models the transformation itself as a GP with the identity function as mean. This transformation $\phi$ in BWGP corresponds to the inverse  of the transformation $\varphi$ in WGP, and can be expressed as 
\begin{equation*}
y(t) =\phi\left( x(t) \right) +\varepsilon _{t},
\end{equation*}
 where $\varepsilon _{t} \sim \mathcal{N}\left( 0,\sigma ^{2}\right)$ and both $x$ and $\phi$ are GPs, that is, 
\begin{align}
x(t) & \sim \mathcal{GP}\left( m(t),k\left( t,\bar{t}\right)\right)\\
\phi\left( f\right)  &\sim \mathcal{GP}\left( f,c\left( f,\bar{f}\right) \right),
\end{align}
where $f$ denotes the input (function) to the warping $\phi$ and $c$ is its covariance kernel. Furthermore, \cite{deep_GP} proposes a deep version of BWGP termed Deep GP (DGP), where the warping function is a composition of multiple GPs. 

Training and inference is intractable both for BWGP and DGP; therefore, both methods rely on a variational approach to perform inference using a sparse representation \cite{titsias2009variational}. Due to their considerable computational complexity, comparisons of the proposed method against BWGP and DGP are beyond the scope of this article, since we focus on expressive warping functions that provide computationally-efficient closed-form formulas for training and prediction. Therefore, experimental validation of the proposed method will be performed against WGP \cite{warped04} only. 

\section{A novel warping for WGPs}
\label{sec:novelWGP}
Inspired by deep architectures, we propose a generative model for non-Gaussian processes by transforming a latent GP through a composition of \emph{elementary functions} $\varphi_i$ with two main objectives: (i) the class of transformations has to be general enough to replicate a wide class of data using few parameters to avoid overfitting and (ii) the approximations required for learning and inference should be minimal so as to maintain high numerical precision and low computational complexity. 

\subsection{Model description} 
\label{sub:model_description}

Let us consider a family of parametric functions $\{\varphi_i\}_{i=1}^d$, $d\in\N$, that are differentiable and invertible with closed-form inverse, hereinafter referred to as \textit{elementary functions}. Then, we can construct warping functions $\varphi (\cdot)$ as a composition of such elementary functions, that is, 
\begin{equation}
	\varphi (\cdot)= \varphi_d(\varphi_{d-1}(\cdots(\varphi_{2}(\varphi_1(\cdot)))\cdots )).
\end{equation}
This construction is motivated by the fact that the inverse and derivatives of function compositions are given by the inverses and derivatives of their component functions. For instance, for a two-elementary-function composition $\varphi(y) =\varphi_{2}(\varphi_{1}(y))=x$, the inverse and the derivative are given respectively by 
\begin{eqnarray*}
	\varphi ^{-1}\left( x\right) &=&\varphi _{1}^{-1}(\varphi _{2}^{-1}(x)) \\
	\frac{d\varphi \left( y\right) }{dy}&=&\frac{d\varphi_{2}\left( \varphi_{1}\left( y\right) \right) }{d\varphi_{1}}\frac{d\varphi_{1}\left( y\right) }{dy}.
\end{eqnarray*}
Notice that this class of warping functions goes one step further compared to WGP in \cite{warped04}: WGP ensures invertibility but then deals with finding the inverse numerically, whereas the compositional warping proposed here ensures both invertibility and closed-form inverses, meaning that the evaluation of the inverse is straightforward.
  
We then propose the compositionally-warped Gaussian process (CWGP) given by $y(t)$ s.t.
\begin{eqnarray*}
	 \varphi(y(t))  &=& 	x(t),\\ 
	 x(t)			&\sim&	\GP(m(t),k(t,\overline{t})),\\  
	 \varphi(\cdot) &=& 	\varphi_d(\cdots(\varphi_2(\varphi_1(\cdot)))\cdots),
\end{eqnarray*}
where $\{\varphi_i\}_{i=1}^d$ are elementary functions.  Additionally, as the inverse of $\varphi$ is known, CWGP can also be interpreted as a generative model that transforms $x(t)$ into $y(t)$ using the transformation $\varphi^{-1}$. For notational clarity we emphasise that $\varphi$ is defined from the non-Gaussian process $y$ to the Gaussian process $x$. 

Finally, we also clarify that the model described above differs radically from the concept of Normalising Flows (NF) \cite{tabak2010,tabak2013,rezende2015variational}. NF focuses on approximating the posterior density of an intractable model, whereas we construct a non-Gaussian generative model directly.

\subsection{Learning: robust, interpretable and efficient} 
\label{sub:sampling}

Learning under CWGP means finding the hyperparameters of the GP $x$ (parameters of the kernel and mean functions denoted by $\theta_x$) in addition to the parameters of the compositional transformation $\varphi$, denoted by $\theta_\varphi$. Thanks to the change of variables theorem, learning these parameters is tractable and can be achieved via minimisation of the negative logarithm of the marginal likelihood (NLL).

\medskip
\noindent\textbf{Robustness.} Just as standard GPs, warped GPs are protected from overfitting, since they directly parametrise a prior distribution over functions and not the specific trajectories of the function. Additionally, recall that the warping considered is component-wise and given by the same scalar-valued map for all the coordinates, thus the warping can be understood as a parametrisation of the marginal histogram. Therefore, the resulting generative model has non-Gaussian marginals with Gaussian copulas, known as \emph{Gaussian copula process} \cite{wilson2010copula}, meaning that in the broad sense of modelling the law of stochastic process, the proposed model is regularised by design.

\medskip
\noindent\textbf{Interpretability.} The NLL is given by 
\begin{align}
\label{eq:NLL}
	\text{NLL} &= -\log p(\y|\theta_x,\theta_\varphi)\\
	 &= \underbrace{\frac{n\log(2\pi)}{2}}_{\text{constant term}} 
	 + \underbrace{\frac{1}{2}\left(\varphi(\y)-\mu_{\x} \right)^{\top}\Sigma_{\x \x}^{-1}\left(\varphi(\y)-\mu_{\x} \right)}_{\text{data-fit  term}}  
	 \nonumber\\ 
	&+ \underbrace{\frac{1}{2} \log \left|\Sigma_{\x \x}\right|}_{\text{kernel-complexity term}} 
	-\quad\underbrace{\sum_{i=1}^{n}\log\left(\frac{d\varphi(y_{i})}{dy}\right)}_{\text{warping-complexity term}},
	\nonumber
\end{align}
where $\mu_\x$ and $\Sigma_{\x\x}$ are the mean and covariance of $\x=\varphi(\y)$. 

Akin to standard GPs, for which the NLL reveals automatic penalty of model complexity, WGP features a \emph{warping-complexity term}. Therefore, the NLL is minimised balancing the Gaussianity of the base GP $x$ via the first three terms in eq.~\eqref{eq:NLL} and the regularity of the warping via the warping-complexity term. Specifically, the first three terms promote solutions such that $||\varphi(\y)-\mu_{\x}||$ is small wrt to the norm induced by $\Sigma_{\x \x}^{-1}$, where the extreme solution is given by $\varphi(\y) = \mu_{\x} = \text{constant } \forall \y,t$, since $\varphi(\y):\y\mapsto x$ and $\mu_{\x}: t\mapsto x$. However, notice that the warping-complexity term $\sum_{i=1}^{n}\log\left(\frac{d\varphi(y_{i})}{dy}\right)$ forces solutions $\varphi(\y)$ that have large derivatives (i.e., which grow steeply), thus ruling out the constant case. These terms offer a clear interpretation of the likelihood function of WGP: the warping-penalty term preserves the data variability by choosing warpings with large derivatives, while the remaining terms ensure that this variability remains as Gaussian as possible. 

\medskip
\noindent\textbf{Computational complexity.} Notice that minimising the NLL does not require the inverse of $\varphi$ but only its log-derivatives, which are known in closed form, therefore, the cost of training CWGP is only dominated by the matrix inversion: $\O(n^3)$ for $n$ observations. Recall that this is the same order of complexity of training standard GPs. Intuitively, learning is then achieved by {transforming} the non-Gaussian observations to then maximise the (Gaussian) probability of the transformed samples wrt to the parameters of (i) the Gaussian distribution and (ii) those of the transformation. Although the complexity of evaluating the NLL is the same for CWGP and standard GPs, the proposed model's NLL might have more local minima due to its larger parameter space. For further details we recommend \cite{ijcnn18}, where multiple local minima are explored with derivative-free and Monte Carlo based optimisation.


\subsection{Closed-form inference} 

Inference follows from a corollary of the change of variables theorem that states that the probability (measure) of a set $E$ under the density of $\y$, is equal to the probability of the image of $E$, $\varphi(E)$, under the density of $\x$. Conditioning on observed data $\y$, we can express the corollary as
\begin{align*}
	\int\limits_{E}p_{y}\left( y|\y\right) dy =\int\limits_{\varphi\left( E \right) }p_{x}\left(x|\y\right) dx=\int\limits_{\varphi\left( E \right) }p_{x}\left(x|\x\right) dx,
\end{align*}%
where the first identity is due to Theorem \ref{thm:CoV} and the second one due to the deterministic relationship $\x=\varphi(\y)$. For different choices of the set $E$ and using the (explicit) inverse transformation $\varphi^{-1}$, we can express the median of $y(t)$ and its $p$-percentile confidence intervals respectively {\bf in closed form} as
\begin{align*}
	\text{median}({y(t)}) &=\varphi ^{-1}\left(\text{median}({x(t)})\right)=\varphi ^{-1}\left(m(t)\right)\\
	I_{y(t)}^p &=\left[ \phi ^{-1}\left(m(t) -z_{p}\sigma(t)\right), \phi ^{-1}\left(m(t) +z_{p}\sigma(t)\right)\right],
\end{align*} 
where $\sigma(t) = \sqrt{k(t,t)}$ is the base GP standard deviation, $z_{p}$ is the $p$-quantile of a standard Gaussian (ex. $z_{0.975} \approx 1.96$) and we used the fact that for a Gaussian $\text{median}(x) = \text{mean}(x)$.

Sampling from CWGP is also straightforward: it is only required to simulate a realisation of the GP and then apply the inverse of the transformation in a coordinate-wise way, that is,
\begin{align*}
 x(\t)&\sim\GP(m(\t),k(\t,\t))\\
 y(\t)& =\varphi^{-1} \left( x(\t) \right). 	
\end{align*}

\subsection{Computational cost of required approximations} 

Relying on the change of variables theorem once again, the expectation of a measurable function $h:\mathcal{Y}\rightarrow \mathbb{R}$ under the non-Gaussian law $p(\y)$ is given by 
\begin{align*}
	\mathbb{E}_{\y}\left[ h\left(\y\right)\right]  =\mathbb{E}_{\x}\left[h\left(\varphi ^{-1}\left( \x\right)\right) \right].
\end{align*}
Additionally, since the distribution of $\x$ is Gaussian, we can efficiently compute the above integral numerically using the Gauss-Hermite quadrature \cite{gausshermite64}, for which  $k$-point approximations are exact when the integrand $h\left(\varphi ^{-1}\left( \cdot\right)\right)$ is a polynomial of order $2k-1$. Choosing $h(y)=y$, we have the approximation of the mean of $y$ given by
\begin{eqnarray}
	\mathbb{E}_{y}\left[ y\right]  &=& \int\limits_{  }\varphi ^{-1}\left( x \right) p_{x}\left(
	x\right) dx \nonumber\\
	 &\approx & \frac{1}{\sqrt{\pi}} \sum\limits_{i=1}^{k} w_{i}\varphi ^{-1}\left( \sqrt{2}\sigma_{x} x_{i} + m_{x} \right), \label{eq:exp_CWGP}
\end{eqnarray}
where the weights $\{w_{i}\}_{i=1}^k$ and locations $\{x_{i}\}_{i=1}^k$ are given by the Gauss-Hermite quadrature method \cite{gausshermite64}. 

Finally, observe that evaluating $\varphi^{-1}$ is required to compute expectations, the median and confidence intervals of the non-Gaussian model. Since for CWGP $\varphi^{-1}$ is known, the cost of evaluating it is $\O(d)$, where $d$ is the number of elementary components of $\varphi$. Therefore, the cost of evaluating $\mathbb{E}_{y}\left[ y\right]$ in eq.~\eqref{eq:exp_CWGP} using the $k$-point Gauss-Hermite quadrature is $\O(kd)$ for CWGP. Conversely, WGP approximates $\varphi^{-1}$ using the Newton-Raphson  method (NRM) \cite{atkinson2008introduction} (with the bisection method to find the initial point), meaning that the cost of evaluating $\mathbb{E}_{y}\left[ y\right]$ for WGP is $\O(kdt)$, where $t$ is the number of iterations of NRM (and bisection). In practice, the explicit expression for $\varphi^{-1}$ is key in computational terms: even using efficient numerical methods, WGP always requires numerical approximations of $\varphi^{-1}$, whereas CWGP is able to evaluate $\varphi^{-1}$ directly.



\begin{table*}[t]
	\centering
	\caption{Elementary transformations: functional forms with derivatives and inverses}
	\label{tab:table_transformations}
	\footnotesize
	\begin{tabular}{l|ccc}
		{Function} &  $\varphi(y)$ &  $\frac{d\varphi(y)}{dy}$ & $\varphi^{-1}(x)$ \\
		\hline 
		Affine	&     $a+by$ & $b$ & $\frac{x-a}{b}$\\
		Logarithmic	&   $\log(y)$ & $y^{-1}$ &  $\exp(x)$ \\
		Arcsinh	&    $a + b\asinh\left(\frac{y - c}{d}\right)$ & $\frac{b}{\sqrt{d^2 + (y-c)^{2}}}$ &  $c + d\sinh\left(\frac{x - a}{b}\right)$ \\
		Box-Cox	&    $\frac{\sgn\left( y\right) \left\vert y\right\vert ^{\lambda }-1}{\lambda }$ & $\left\vert y\right\vert^{\lambda -1}$ & $\sgn\left(\lambda x+1\right) \left\vert	\lambda x+1\right\vert ^{\frac{1}{\lambda }}$ \\
		Sinh-Arcsinh	& $\sinh\left(b\asinh(y) - a\right)$ & $\frac{b\cosh\left(b\asinh(y) - a\right)}{\sqrt{1 + y^{2}}}$ & $\sinh\left(\frac{1}{b}\left(\asinh(x)+a\right)\right)$ \\
	\end{tabular}
\end{table*}

\begin{figure*}[t]
	\centering
	\includegraphics[width=.8\textwidth]{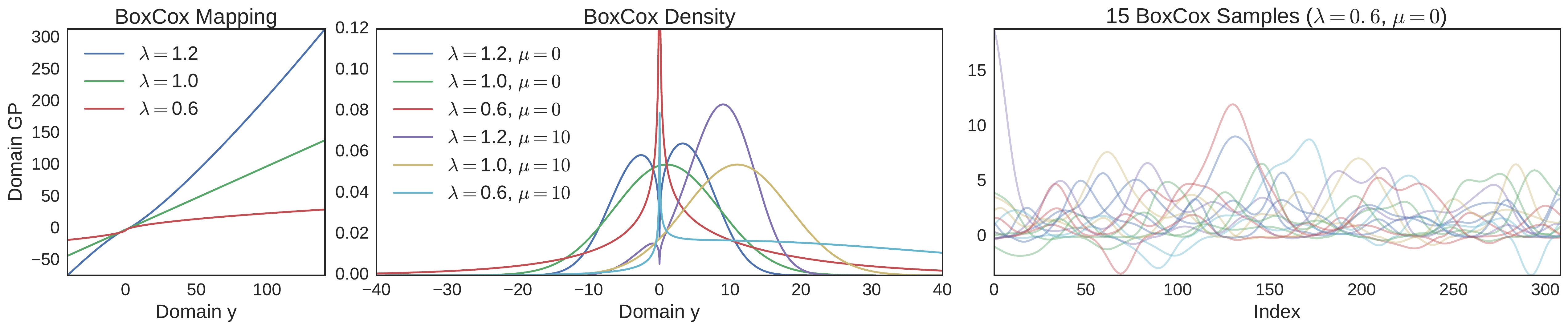}\\
	\includegraphics[width=.8\textwidth]{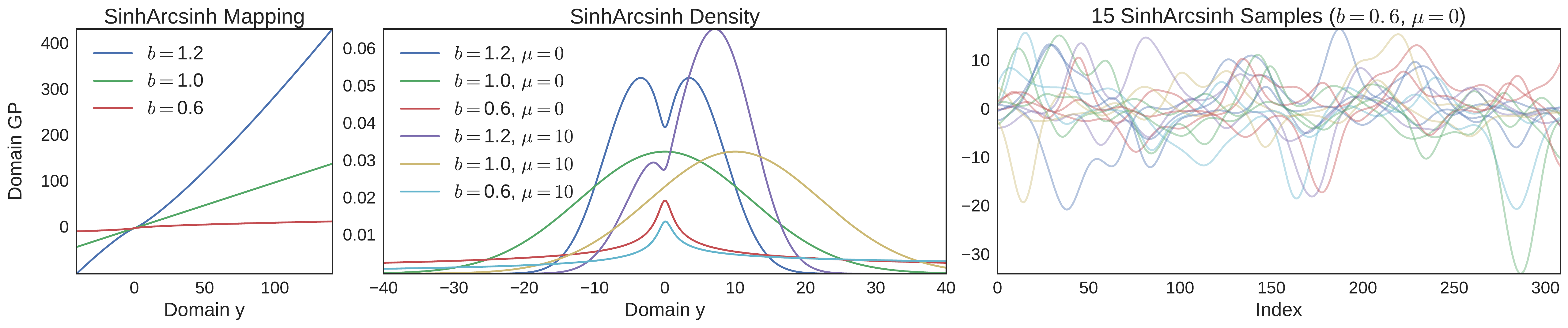} 

	\caption{Proposed Box-Cox and SinhArcsinh elementary transformations. \textbf{Left:} transformations (or warpings), \textbf{centre:} induced marginal densities, \textbf{right:} samples of the warped GP, \textbf{top:}   Box-Cox transformation in eq.\eqref{eq:box-cox-trans}, and \textbf{bottom:} SinhArcsinh transformation in eq.~\eqref{eq:sinharcsinh-trans}. For all plots, $\mu$ denotes the mean of the base GP $x$.}
	\label{fig:tgp_processes}
\end{figure*}

\section{Elementary transformations} 
\label{sec:transformations}

As a companion to the CWGP proposed in the previous section, we now present a set of elementary transformations with explicit inverse and derivative to be used as building blocks of CWGP's compositional transformation. Furthermore, for consistency with Theorem \ref{thm:CoV}, we present the transformations from the non-Gaussian process $y$ to the GP $x$. Table \ref{tab:table_transformations} gives a summary of these transformations together with their inverses and derivatives, and Fig. \ref{fig:tgp_processes} shows two different families of transformations together with their induced marginal densities and sample trajectories.

\subsection{Affine transformation}
 
The affine transformation is given by 
\begin{equation}
\label{eq:affine-trans}
\varphi_\text{affine} (y) = a+by,\ \ a,b\in\R,	
\end{equation}
and is referred to as \emph{shift} when $b=1$ and as \emph{scale} when $a=0$. The affine transformation does not provide enhanced modelling ability over standard GPs, since an affine-transformed GP is still a GP with a shifted mean and scaled variance. However, the affine warping will be composed with other elementary functions to produce expressive transformations. 

\subsection{Box-Cox transformations}
\label{sec:BC_trans}
A standard strategy in Statistics to transform non-Gaussian positive observations into \textit{closer-to-Gaussian} ones is to apply the logarithmic function $\varphi_{\log} (y)= \log(y)$; this is the case for positive-valued heavy-tailed stochastic processes \cite{aitchison1976lognormal} such as word counts \cite{twitter}. Notice that with the logarithmic transformation, both the mean $m_x$ and variance $\sigma_x^2$ of the original GP $x$ affect all moments of the transformed process $y$. Explicitly, the $n\text{-th}$ moment of $y$ is given by 
\begin{equation}
	\mathbb{E}_{y}\left[ y^{n}\right] =\exp \left( nm_{x}+\tfrac{1}{2}n^{2}\sigma_{x}\right),
\end{equation} meaning that a heavy-tailed distribution for $y$ is obtained through only modifying the mean and variance of the original process $x$.

A generalisation of the logarithmic transformation is the Box-Cox transformation \cite{boxcox,sakia1992box}, a single-parameter power function given by
\begin{eqnarray}
\label{eq:box-cox-trans}
	\varphi_{\lambda}\left( y\right) &=&\frac{\sgn\left( y\right) \left\vert
		y\right\vert ^{\lambda }-1}{\lambda }, \ \lambda \in \mathbb{R}^{+}_{0},
\end{eqnarray}
where $\varphi_{\lambda}$ becomes a power function for $\lambda >0$, an affine transformation for $\lambda=1$, and the logarithmic transformation for $\lambda =0$  since $\lim\limits_{\lambda \rightarrow 0 }\varphi _{\lambda}(y) =\log(y)$. 

The Box-Cox transformation has two key properties: Firstly, its mode is known \cite{powernormal} to be 
\begin{align*}
\text{mode}_y=\left[ \frac{1}{2}\left( 1+\lambda m_x+\sqrt{\left( 1+\lambda m_x\right)^{2}+4\sigma_x^2\lambda \left( \lambda -1\right) }\right) \right] ^{\frac{1}{\lambda }},
\end{align*}
where $m_x$ and $\sigma_x^2$ are the mean and variance of the GP $x$ respectively. This is particularly useful for skewed distributions where the mode is usually considered as point estimate rather than the mean or the median. Secondly, the computation of moments using numerical methods, e.g., the Gauss-Hermite quadrature \cite{gausshermite64}, can be performed with high precision due to the polynomial nature of the Box-Cox transformation. Fig.~\ref{fig:tgp_processes} (top) shows different Box-Cox transformations with their induced marginal densities. 

\subsection{Hyperbolic transformations}
The distribution resulting from passing a $\cN(0,1)$ random variable through the inverse hyperbolic sine transformation 
\begin{align}
\label{eq:arcsinh-trans}
	\varphi_\text{arcsinh}\left( y\right) = a + b\arcsinh\left(\frac{y - c}{d}\right),
\end{align}
where $a, c \in \mathbb{R}$ and $b,d \in \mathbb{R}^{+}$, is known as the Johnson's SU-distribution \cite{johnsonsu} and has closed-form expressions for the mean and variance, given respectively by 
\begin{align*}
\mu_\text{SU} &= c - d\exp\left(\frac{b^{-2}}{2}\right)\sinh\left(\frac{a}{b}\right)\\
\sigma_\text{SU}^{2} &= \frac{d^{2}}{2}\left[\exp\left(b^{-2}\right)-1\right]\left[\exp\left(b^{-2}\right)\cosh\left(\frac{2a}{b}\right) + 1\right],
\end{align*}
and also for the skewness and kurtosis \cite{johnsonsu}.

Another transformation based on hyperbolic functions is the  Sinh-Arcsinh \cite{Sinharcsinh}, where arcsinh, affine and sinh are composed together, that is,
\begin{equation}
\label{eq:sinharcsinh-trans}
\varphi_\text{SinhArcsinh}\left( y\right) = \sinh\left(b\arcsinh(y) - a\right),
\end{equation}
where $a,b\in\R$. This distribution admits explicit expressions for all moments of $y$, using the modified Bessel function, and it induces a distribution where the third and fourth moments can be controlled via parameters $a$ and $b$. This distribution is symmetric if $a=0$; positively-skewed (cf. negatively-skewed)  if $a > 0$ (cf. $a < 0$); mesokurtic if $b=1$; and leptokurtic (cf. platykurtic) is $b > 1$ (cf. $b < 1$). Additionally, for this distribution we have that  $0 < |\text{mode}(y)| < \sinh(|a|/b)$ and $\sgn(\text{mode}(y))=\sgn(a)$. 

Fig.~\ref{fig:tgp_processes} (bottom) shows the Sinh-Arcsinh transformations with the induced marginals and samples for a skewness parameter set to $a=0$ and different values of the kurtosis parameter $b$. Observe that the mean of the base GP, $\mu$, can also change the skewness of the induced marginal distribution.

\section{How to choose the elementary transformations?}
\label{sec:choice_trans}

As in the vast majority of deep structures, the number of layers and the type of neurons are defined by experts or by trial and error, where interpretability is a desired property \cite{representation_NN}. This is also the case when choosing the kernel in support vector machines or Gaussian processes (as studied in \cite{duvenaud2013structure}). Recall that in standard mixture models (such as WGP) the user only defines the number of components, whereas within the proposed CWGP one also needs to choose the types and ordering of the components (in our case, elementary functions). This section provides guidance on the choice of the elementary transformations under two scenarios, the first one being the case when expert knowledge about the data is available. In the second scenario, that is, when no prior knowledge of the data is available, we show that CWGP can be implemented by concatenating multiple instances of a particular sequence of elementary transformations (referred to as the \emph{SinhArcsinh-Affine} layer); we will see that this construction has appealing experimental performance. This way, CWGP can be regarded as a black-box where, akin to deep structures,  the user only needs to choose the number of layers. We illustrate this concept based on the NLL (Sec.~\ref{sub:sparse_trans}) and via a toy example (Sec.~\ref{sub:replicating_wgp}), as well as its robustness to overfitting and through a real-world data in Sec.~\ref{sub:sunspots}.

\subsection{When prior knowledge of the data is available}

As mentioned in Sec.~\ref{sec:BC_trans}, when the data are strictly positive a standard practice is to apply the logarithmic transformation. Critically, if the data are known to be lower-bounded by an unknown quantity, one can compose the logarithmic transformation with the shift transformation in eq.~\eqref{eq:affine-trans} in order to find the shift parameter during training. An upper bound to the data can be found in an analogous way by replacing the shift by an affine transformation, thus allowing for a negative scaling. In this sense, composing two affine-logarithmic transformations enables us to find the upper and lower bounds simultaneously. 

To further relax the strict (lower) bound condition of the logarithmic transformation to a more permissive one, we can also replace the logarithm by the Box-Cox transformation in eq.~\eqref{eq:box-cox-trans}, where the permissiveness of the bound is controlled by the parameter $\lambda$. Additionally, if the data is such that their range is not bounded but rather have a large dispersion, then the data follows a heavy-tailed distribution. This phenomenon can be modelled using the Arcsinh or Sinh-Arcsinh transformations in eqs.~\eqref{eq:arcsinh-trans} and \eqref{eq:sinharcsinh-trans} respectively, since such transformations allow to control the mean and variance of the distribution, as well as its asymmetry and kurtosis. All these transformations can be composed with one another to construct more complex distributions, as in the case of multimodal distributions.

\subsection{Sparse compositional transformations} 
\label{sub:sparse_trans}

As in any model that involves choosing a finite order (such as layers, neurons, components), it is required that the addition of more elementary functions in CWGP results in a monotonically-increasing performance. In particular, if one considers an unnecessarily-large number of elementary transformations, it is desired that some of these transformations \emph{revert} to the identity function (and thus can be removed). If, after training, some of the transformations considered revert to the identity, we will say that the compositional transformation is sparse.  

When insight into the statistical properties of the data is scarce, or even non-existent, a recommended procedure is to sequentially add transformations that can revert to the identity when needed. Notice that if a transformation is not able to improve performance and at the same time is able to revert to the identity, it will indeed revert to the identity. This can be justified based on the NLL in eq.~\eqref{eq:NLL}: where the data-fit term remains invariant and the warping-complexity term contributes to a lower NLL. Additionally, one can always choose a prior distribution over the warping parameters to further promote warpings that are close to the identity. Lastly, recall that from the proposed transformations, the Box-Cox, the Sinh-Arcsinh and the affine transformations can revert to the identity, therefore, under limited knowledge about the underlying properties of the data, we recommend to add these components iteratively until the performance of the model reaches a plateau. We next implement this concept based only on the Sinh-Arcsinh and affine transformations on synthetic data and, in Section \ref{sub:sunspots}, on real-world data.

\subsection{Structure discovery via deep compositional transformations} 
\label{sub:replicating_wgp}

For the cases when expert knowledge about the nature of the data is scarce, the proposed CWGP can be implemented just concatenating multiple instances of the proposed elementary transformations, this procedure is usual and widely accepted in general deep architectures \cite{bengio2009learning,Goodfellow-et-al-2016,DL_overview}. To illustrate this, let us first define the composition of a Sinh-Arcsinh and Affine transformations, in eqs.~\eqref{eq:arcsinh-trans} and \eqref{eq:affine-trans} respectively, as the \emph{SAL layer}\footnote{The acronym SAL comes from SinhArcsinh and Affine, where the use of ``L'' stems from ``linear''. This nomenclature has been chosen to be consistent with the experimental part in next section. } given by
\begin{equation}
\label{eq:layer}
	l(y)=a + b\sinh(c\arcsinh(y)-d),
\end{equation}
where $a,b,c,d\in\R$ are the only four parameters of the so-defined layer. We next show that, by only concatenating SAL layers, we can replicate the sum-of-hyperbolic-tangent warping implemented by WGP \cite{warped04}, in eq.~\eqref{eq:wgp}. The reason to assess the proposed model in the approximation of the WGP is that the sum of hyperbolic tangents is known to be \emph{universal}, meaning that it can approximate continuous functions to any desired degree of accuracy in a closed interval. 

\begin{figure}[t]
	\centering
	\includegraphics[width=0.4\textwidth]{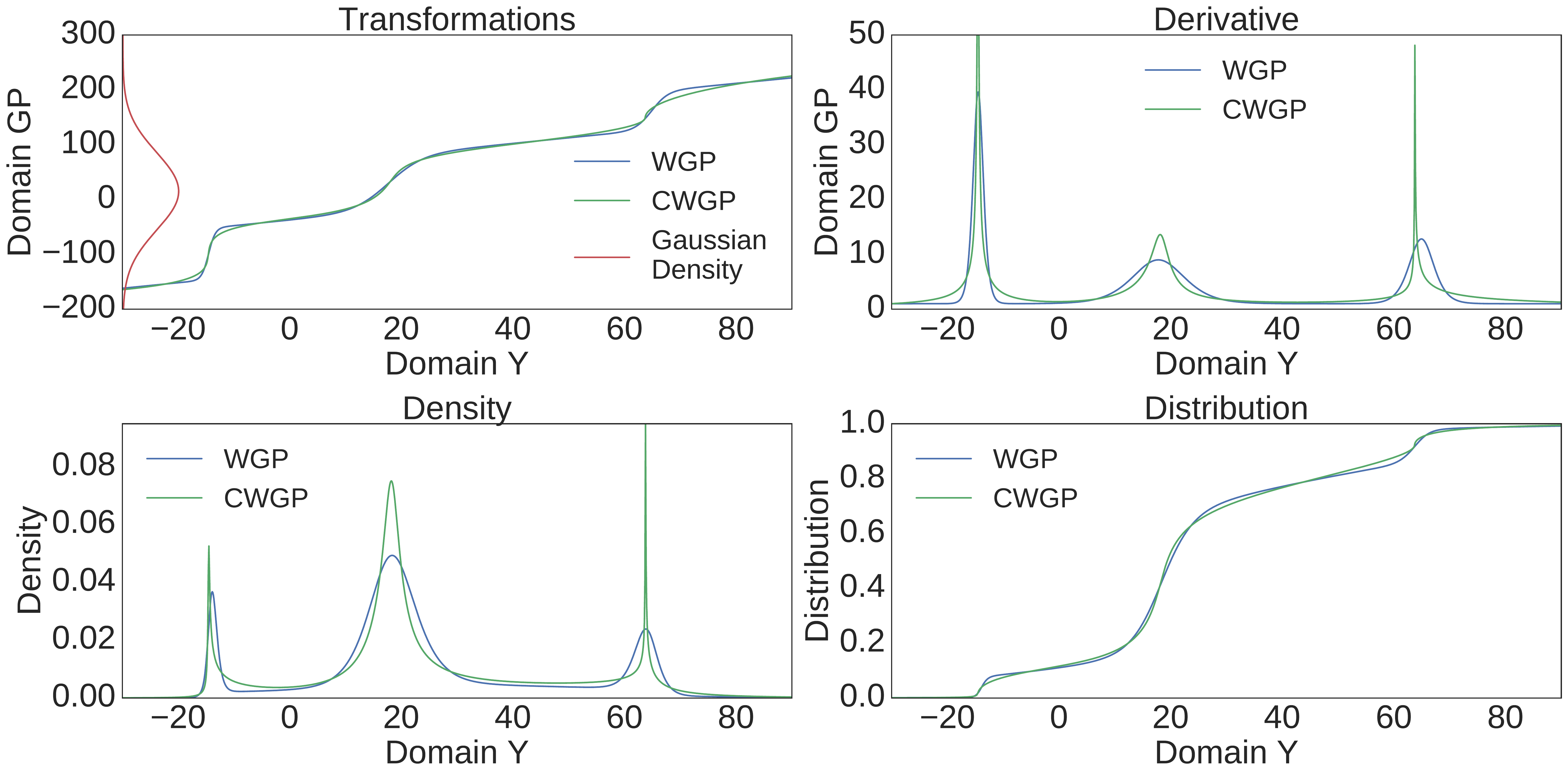}\\
	\caption{Approximation of a WGP warping (sum of three hyperbolic tangents, blue) using the proposed compositional method (three SAL layers, green).}
	\label{fig:cwgp_flexible}
\end{figure}

With the aim of gaining an intuitive understanding about the modelling ability of the compositional approach, the first illustrative example is to train a three-SAL-layer compositional transformation, via least squares, to replicate a mixture of three hyperbolic tangents. Fig.~\ref{fig:cwgp_flexible} shows the transformations, derivatives, densities and distributions of the ground truth (WGP, blue) and the three-SAL-layer compositional approximation (CWGP, green). Observe the point-wise similarity of the warpings and that the probability mass is concentrated around the three common modes in the domain of $y$.

\begin{figure}[t]
	\centering
	\includegraphics[width=0.4\textwidth]{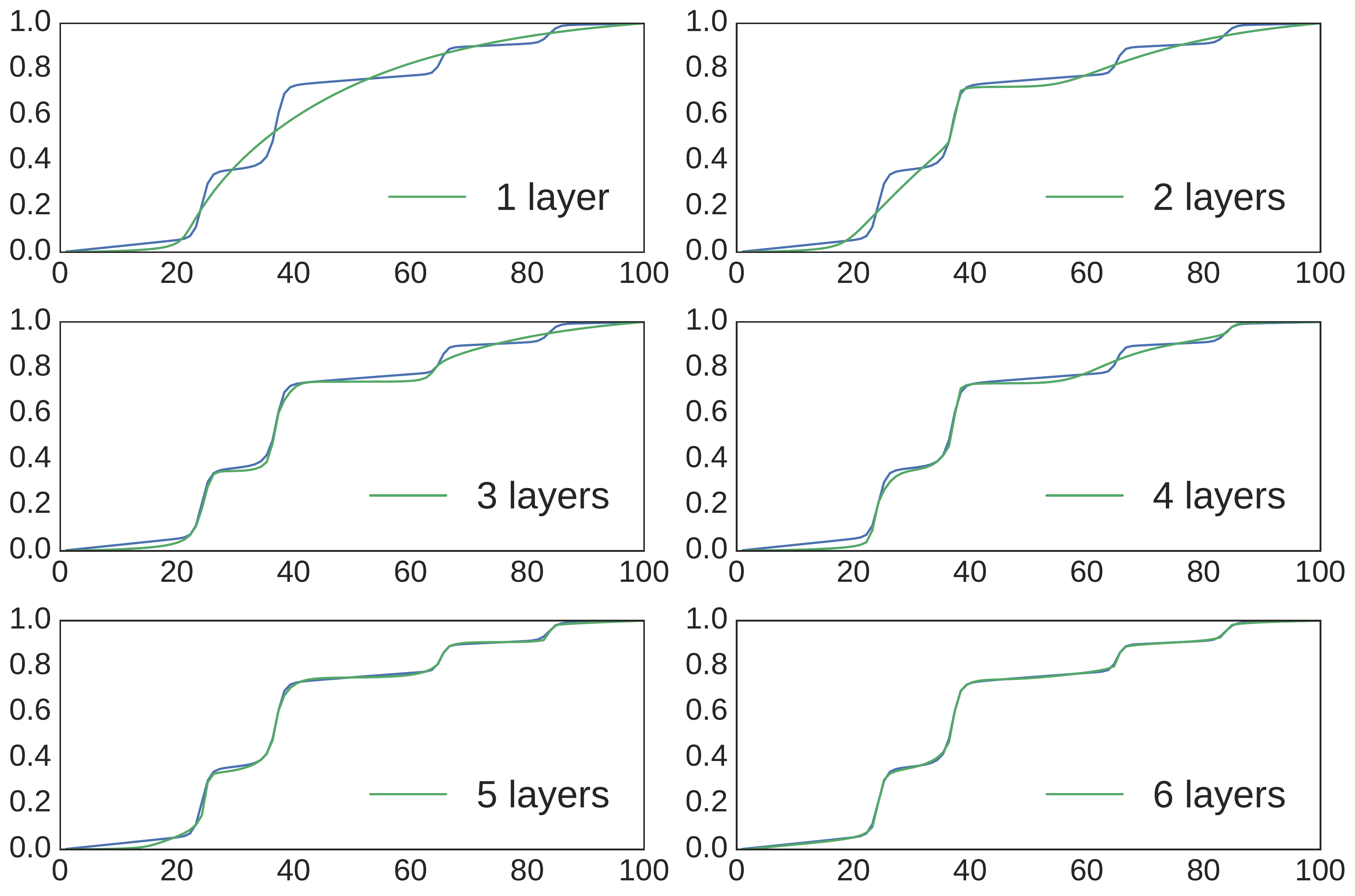}\\
	\caption{CWGP approximation of the distribution of a WGP with five hyperbolic tangents: Ground truth (blue) and CWGP approximations (green) using a variable number of SAL layers in eq.~\eqref{eq:layer}.}
	\label{fig:cwgp_blackbox1}
\end{figure}

\begin{figure}[t]
	\centering
	\includegraphics[width=0.4\textwidth]{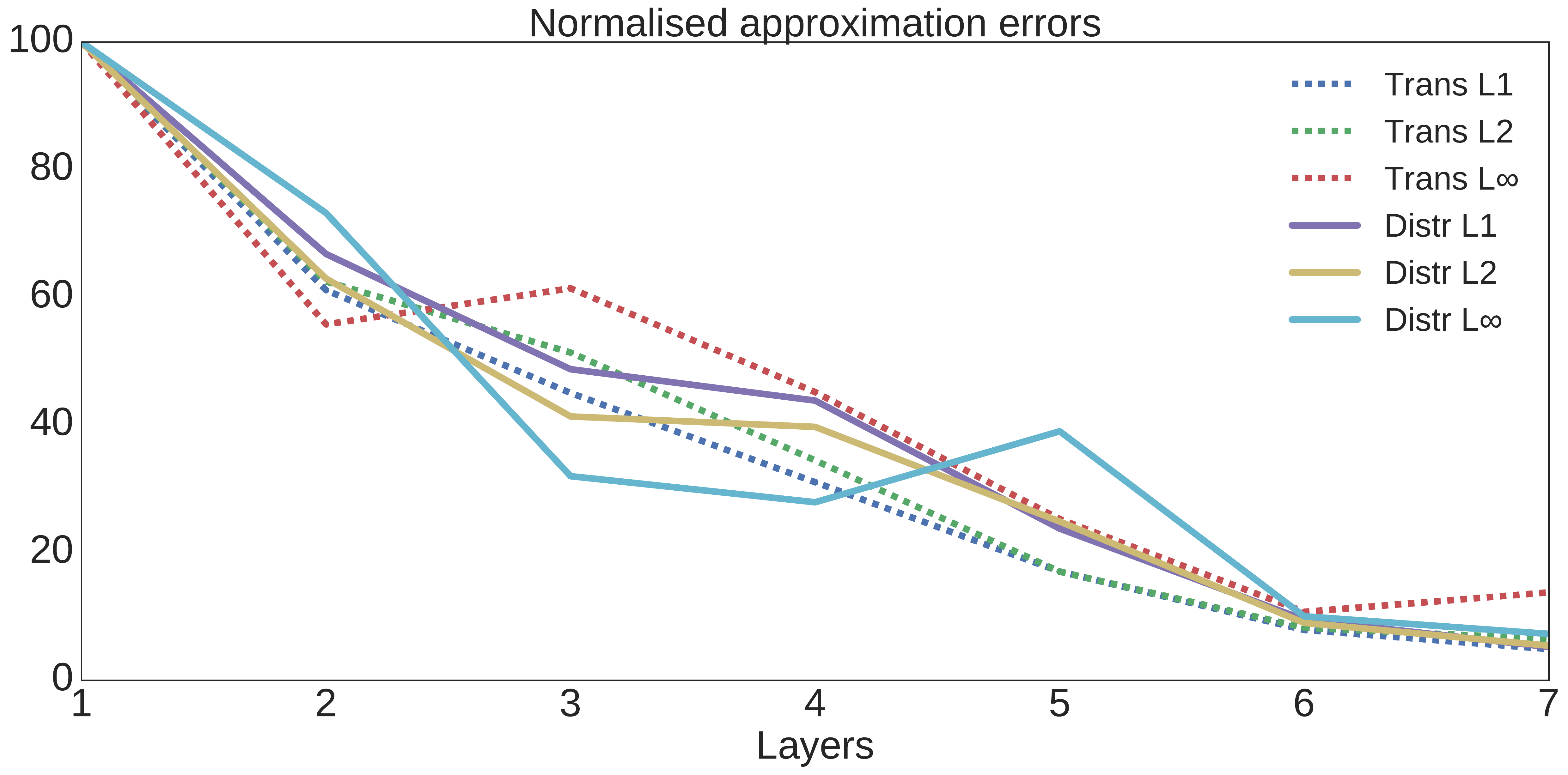}\\
	\caption{Representation of error measures in Table \ref{tab:table_blackbox} normalised wrt to the error of the single-layer case.}
	\label{fig:cwgp_blackbox2}
\end{figure}

Regarding the expressiveness of the proposed compositional approach as a function of the number of considered SAL layers, Fig.~\ref{fig:cwgp_blackbox1} shows the induced distributions for a five-hyperbolic-tangent WGP warping (blue) and those of the compositional approximations using one to six layers (green) fitted by least squares. Notice how the distributions learnt by the compositional transformation becomes indistinguishable from the ground truth as the number of SAL layers increases. Table \ref{tab:table_blackbox} reports the approximation errors both for the transformation and the resulting (warped) distribution, using the $L_1, L_2$ and $L_\infty$ norms given respectively by
\begin{alignat*}{3}
 e_1 &=||f_{\text{SoT}}-f_{\text{CT}}||_1 &&=\int_\R|f_{\text{SoT}}(x)-f_{\text{CT}}(x)|dx\\
  e_2 &=||f_{\text{SoT}}-f_{\text{CT}}||_2 &&=\sqrt{\int_\R|f_{\text{SoT}}(x)-f_{\text{CT}}(x)|^2dx}\\
   e_\infty &=||f_{\text{SoT}}-f_{\text{CT}}||_\infty &&=\sup_{x\in\R}|f_{\text{SoT}}(x)-f_{\text{CT}}(x)|,
\end{alignat*}
where $f_{\text{SoT}}$ denotes de transformation (or distribution) of WGP's \emph{sum of hyperbolic tangents}, and $f_{\text{CT}}$ those of the proposed \emph{compositional transformation}. Fig.~\ref{fig:cwgp_blackbox2} also shows the above error measures normalised wrt the to the single-layer case---observe the monotonic performance of the approximation as the number of SAL layers increases.

\setlength{\tabcolsep}{3pt}
\begin{table}[t]
	{\small
		\caption{Black-box approximation of a WGP warping with five hyperbolic tangents: $L_1, L_2$ and $L_\infty$ error measures for transformations and induced distributions for different number of layers}
		\label{tab:table_blackbox} 
	}
	\centering
	\footnotesize
	\begin{tabular}{c|ccc|ccc}

	\toprule
	layers &    Tran L1 &   Tran L2 &  Tran L$\infty$ & Dist L1 & Dist L2 & Dist L$\infty$ \\
	\midrule
	1      & 1878.2 & 243.11  & 60.57 &  3.342 &  0.464 &  0.147 \\
	2      & 1147.7 & 151.86  & 33.77 &  2.232 &  0.292 &  0.107 \\
	3      &  845.14 & 124.80 & 37.19 &  1.628 &  0.192 &  0.047 \\
	4      &  582.53 &  83.71 & 27.34 &  1.464 &  0.184 &  0.041 \\
	5      &  319.64 &  41.33 & 15.28 &  0.793 &  0.115 &  0.057 \\
	6      &  147.78 &  19.95 & 6.48 &  0.316 &  0.042 &  0.015 \\
	7      &   91.64 &  15.71 &  8.32 &  0.174 &  0.025 &  0.011 \\
	\bottomrule
	\end{tabular}
\end{table}

\section{Experimental validation}
\label{sec:experiments}

We evaluated CWGP experimentally in three real-world scenarios. The first one has an illustrative purpose and demonstrates the robustness of CWGP wrt the number of chosen elementary functions using an astronomical time series. The second experiment validates the ability of the proposed CWGP to identify key statistical properties of a real-world financial time series. Lastly, the third experiment tests CWGP on the three datasets originally used in \cite{warped04,bayesianwarped12}, where our aim is to assess the proposed model in terms of predictive performance and experimental computational complexity.

We compared the proposed CWGP against GP and WGP only, and left BWGP and DGP out of this study due a number of reasons. First, our aim is to propose a computationally efficient method with exact inference and minimal numerical approximations for prediction, BWGP and DGP fall well outside this aim due to their intractable inference. Second, both BWGP and DGP rely on variational inference (VI) methods, therefore, the performance of BWGP/DGP depends on the considered approximation; consequently, comparison using off-the-shelf VI methods might be misleading. Third, according to \cite{bayesianwarped12}, the standard WGP performed better that BWGP in five out of six performance indices for the same datasets; as we consider those datasets in Sec.~\ref{sub:real-world}, we are also indirectly comparing against BWGP. Finally, we believe that the availability of an invertible warping is key for interpreting the relationship between the base GP and the transformed (non-Gaussian) process, as this leads to discovering statistical properties of the data; this is a advantage of the CWGP that neither BWGP nor DGP can provide.

We next define performance indices to be used in our experimental evaluation to then proceed to the simulations.

\subsection{Performance indices} 
\label{sub:indices}

For consistency with the existing literature on warped GPs \cite{warped04,bayesianwarped12} and  to give a thorough evaluation of the model proposed, we considered four performance indices: the negative log-likelihood (NLL), the root mean squared error (RMSE), the mean absolute error (MAE), and the negative log predictive distribution (NLPD). These indices are described below and should be interpreted as \emph{the lower the better}.

Firstly, the NLL in eq.~\eqref{eq:NLL} is a measure of the {probability} of the observed data under the chosen model. Model selection and fitting will be achieved by minimising the NLL wrt to the model parameters and hyperparameters. 

Let us now denote a test set $\{y_i\}_{i=1}^n$ and the reported predictive means $\{y^*_i\}_{i=1}^n$, and define the RMSE and the MAE respectively by 
\begin{align}
	\text{RMSE} &= \left(\frac{1}{n}\sum_{i=1}^{n}(y_{i} - y_{i}^{*})^2\right)^\frac{1}{2} \label{eq:RMSE}\\
	\text{MAE} &= \frac{1}{n}\sum_{i=1}^{n}|y_{i} - y_{i}^{*}| \label{eq:MAE}
\end{align}
these two indices are representative of point prediction errors.

Lastly, the NLPD, a measure of the (not necessarily Gaussian) distribution prediction error is defined by
\begin{align}
	\label{eq:NLPD}
	\text{NLPD} &=-\frac{1}{n}\sum_{i=1}^{n}\log(p_{i}(y{}_{i})),
\end{align}
where  $\{p_i(\cdot)\}_{i=1}^n$ are the learnt predictive densities. 

In addition to the above performance indices, the models considered are also evaluated in terms of their training and evaluation times in the second set of experiments.

\subsection{Testing for robustness with the Sunspots time series} 
\label{sub:sunspots}

The aim of this example is to show that adding more elementary functions to the CWGP only improves performance and does not overfit to the training set. Using the Sunspot time series \cite{sunspots} corresponding to the yearly number of sunspots between 1700 and 2008 (309 data points), we randomly selected half of data between 1700 and 1961 (131 observations) as training set. The remaining data points were used for evaluation: the data between 1700 and 1961 not used for training (131 test points) were used for a \emph{reconstruction} experiment, whereas the data after 1961 (47 test points) were used for a \emph{forecasting} experiment. 

As the Sunspot series is positive valued and semiperiodic, we used the CWGP with a 2-component spectral mixture (SM) kernel \cite{Wilson:2013,MOSM} and different quantities of Box-Cox and Sinh-Arcsinh elementary functions. Each model was trained minimising the NLL in eq.~\eqref{eq:NLL} using both the Broyden–Fletcher–Goldfarb–Shanno algorithm (BFGS) \cite{wright1999numerical} and the derivative-free global optimisation Powell \cite{powell1964efficient}; this choice was due to the large number of local minima that characterises spectral-based kernels \cite{bnse,nips15,npr_15b} and follows \cite{ijcnn18}.

\begin{figure}[ht]
	\centering
	\includegraphics[width=0.4\textwidth]{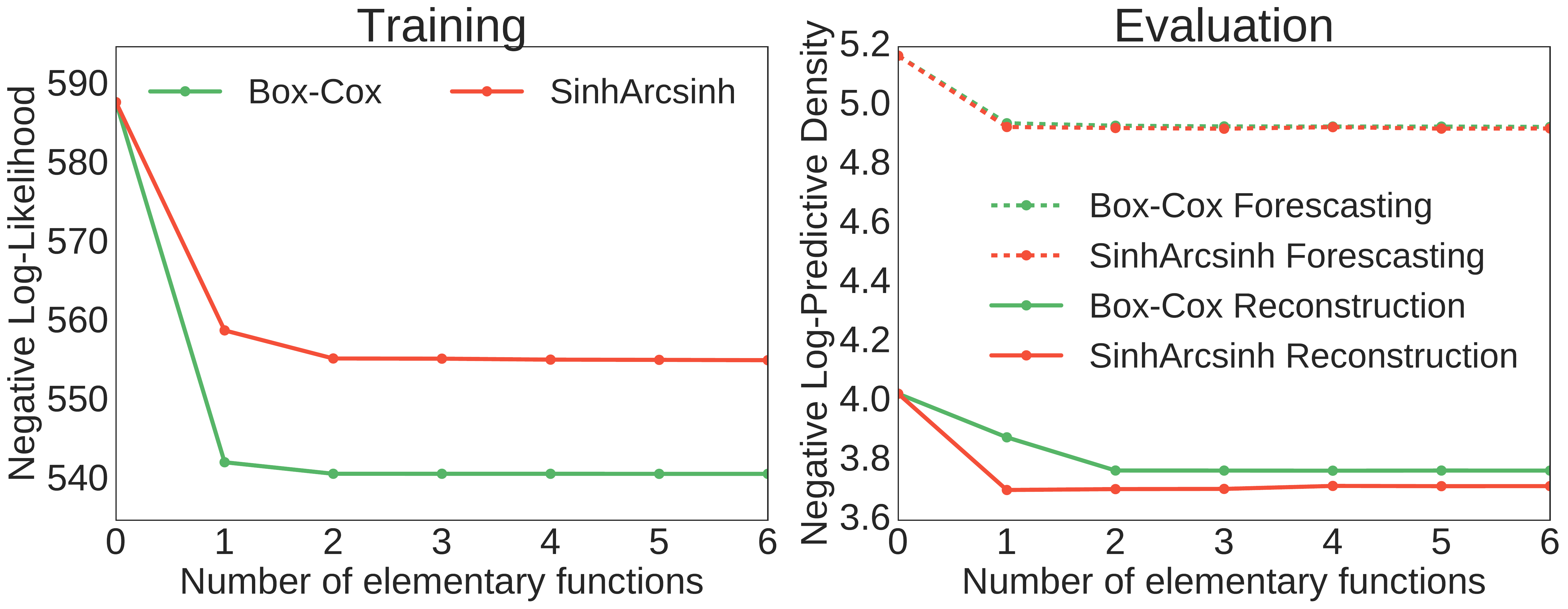}
	\caption{Training (left, NLL) and evaluation (right, NLPD) performance of Box-Cox and Sinh-ArcSinh compositions as a function of the number of elementary transformations. Evaluation is assessed over the reconstruction and forecasting experiments.}
	\label{fig:sunspots_perf}
\end{figure}

\begin{figure}[ht]
	\centering
	\includegraphics[width=0.4\textwidth]{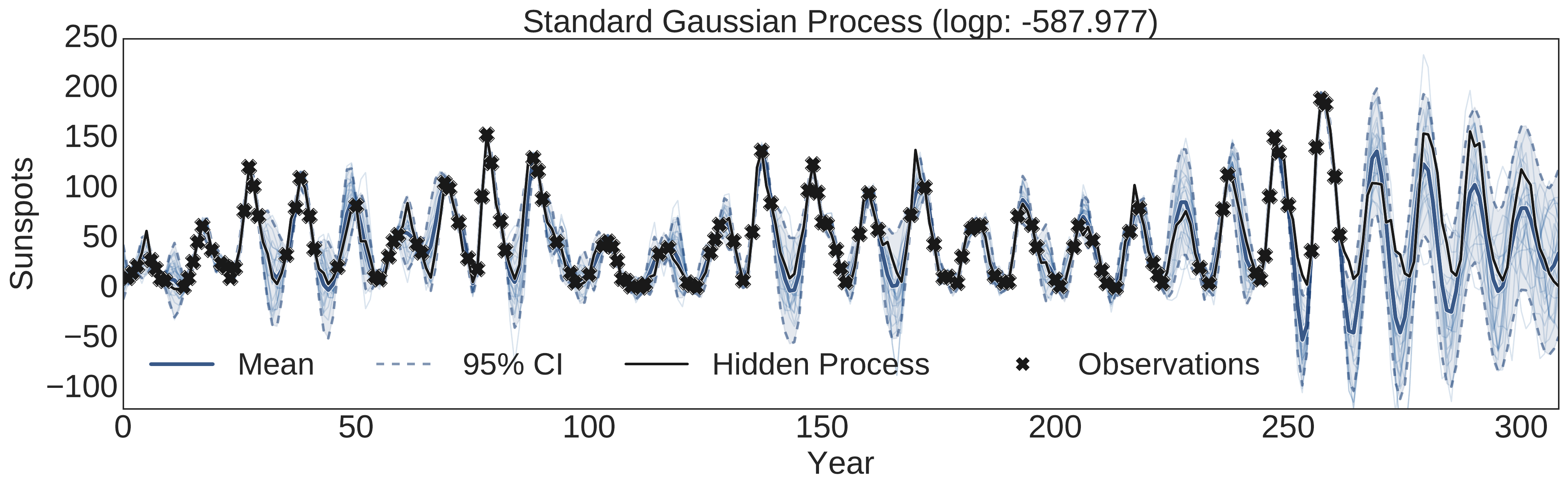}\\
	\includegraphics[width=0.4\textwidth]{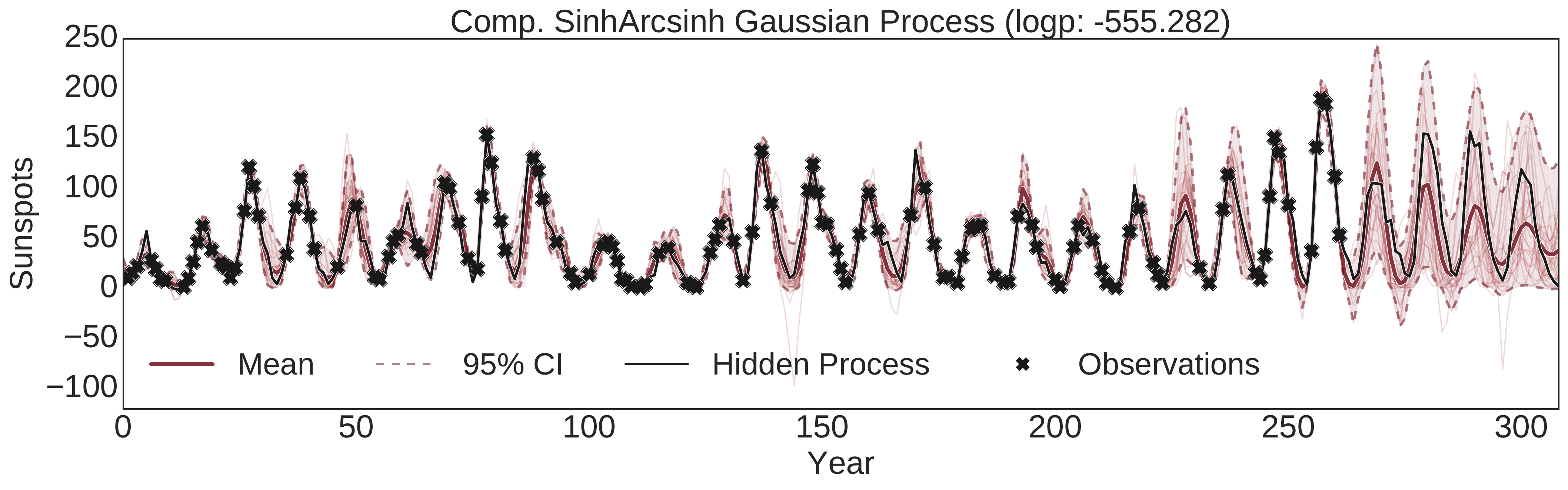}\\
	\includegraphics[width=0.4\textwidth]{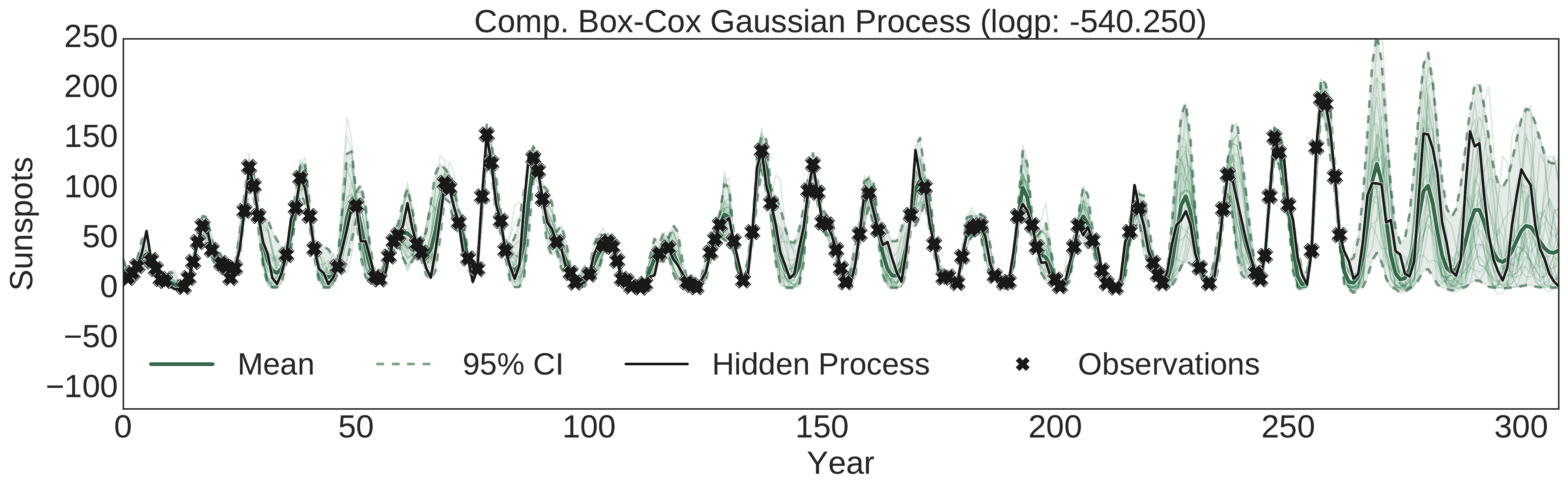}
	\caption{Posterior distribution over sunspots trajectories: GP (top), 6-component SinhArcsinh (middle), and 6-component Box-Cox (bottom). Notice the tighter error bars of the CWGP models and the skewed marginal posteriors that are concentrated on positive values.}
	\label{fig:sunspots}
\end{figure}

Fig.~\ref{fig:sunspots_perf} shows the performance (NLL and NLPD) as a function of the number of elementary functions of both models, where zero elementary functions means standard GP. Notice how these experiments confirm the robustness-to-overfitting ability of the CWGP, where despite the unnecessary addition of elementary functions, the validation performance does not degrade---even for forecasting. Also, Fig.~\ref{fig:sunspots} shows the trained models with zero elementary functions (standard GP, top) and 6 elementary functions for the Sinh-ArcSinh (middle) and Box-Cox (bottom) compositions.

\subsection{Learning a macroeconomic time series} 
\label{sub:adjusting_prior}

We then implemented CWGP alongside a standard GP to learn the quarterly average \emph{3-Month Treasury Bill: Secondary Market Rate} \cite{tb3ms} between the first quarter of 1959 and the third quarter of 2009, that is, 203 observations. We knew beforehand that this macroeconomic signal is the price of U.S. government risk-free bonds, which cannot take negative values and can have large positive deviations. Therefore, we implemented CWGP with a warping consisting of one affine and one Box-Cox elementary transformations in eqs.~\eqref{eq:affine-trans} and \eqref{eq:box-cox-trans} respectively. This experiment reveals the ability of CWGP to identify the complex statistical properties of the data---where the standard GP fails. 

\begin{figure}[ht]
	\centering
	\includegraphics[width=0.4\textwidth]{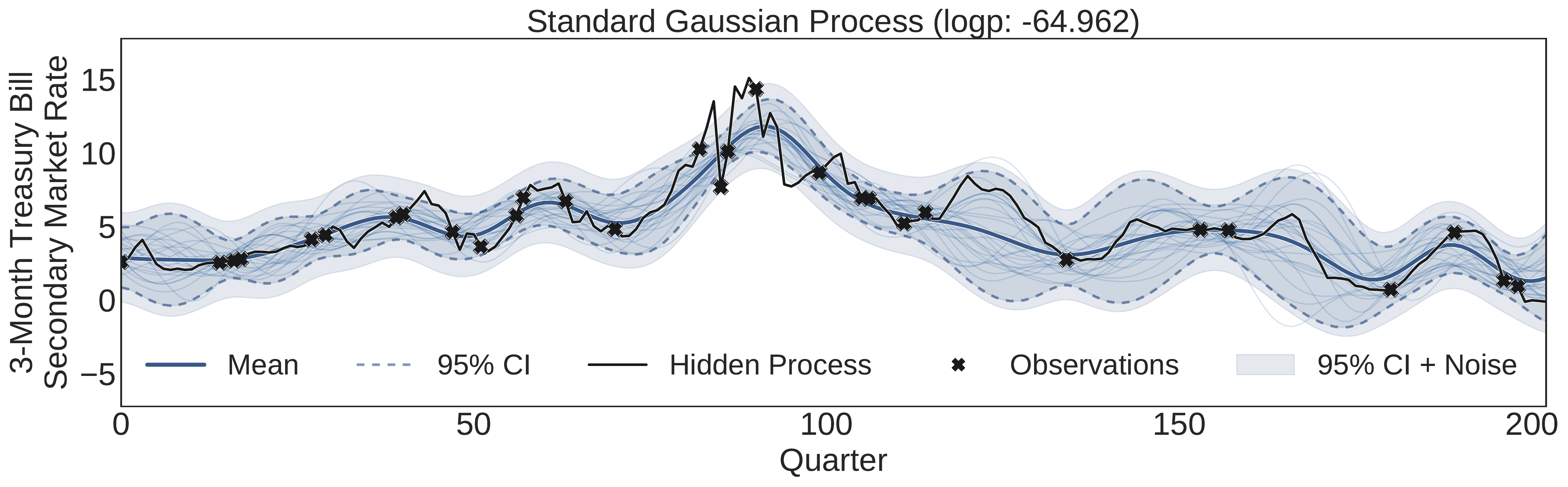}
	\includegraphics[width=0.4\textwidth]{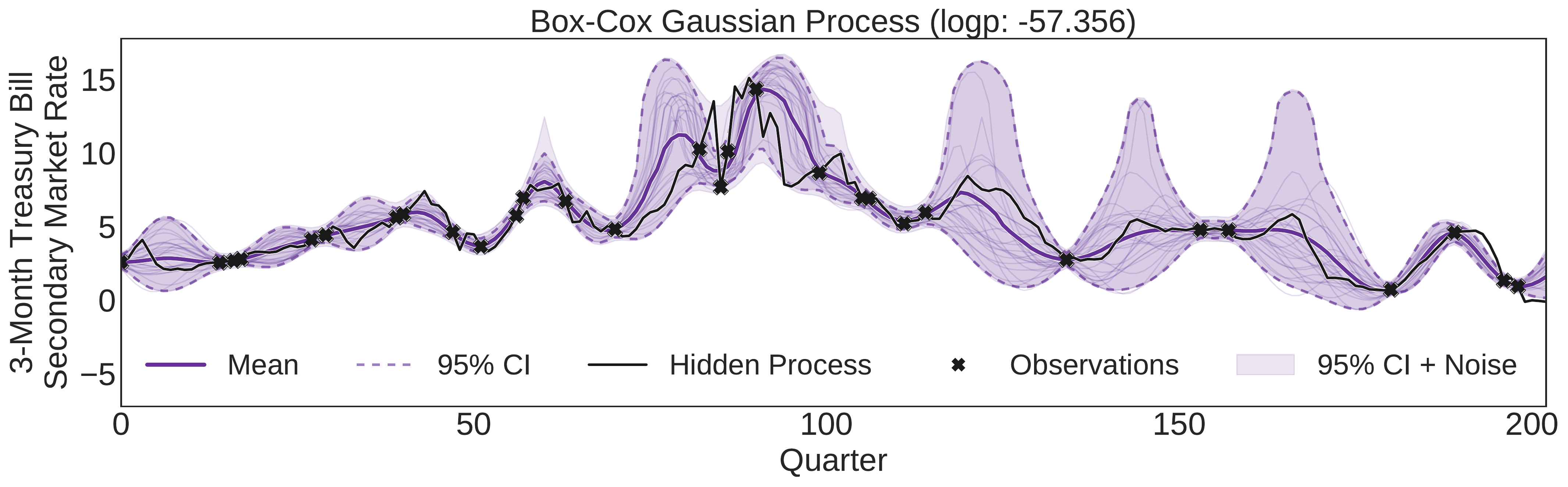}
	\caption{Posterior distribution of the \emph{Quarterly Average 3-Month Treasury Bill: Secondary Market Rate} between 1959 and 2009 using 40 observations (203 datapoints in total). \textbf{Top}: Standard GP. \textbf{Bottom}: Proposed CWGP. Both models used a constant mean and a SE kernel. The CWGP warping comprised an Affine and a Sinh-arcsinh transformation.}
	\label{fig:gp_cwgp_processes}
\end{figure}

Fig.~\ref{fig:gp_cwgp_processes} shows both GP (top) and CWGP (bottom) posterior distributions with only 40 observations for the time series, together with their means, error bars and sample trajectories, while Table \ref{tab:table_macroeconomic} shows the performance metrics. Notice the evident non-Gaussianity of the posterior revealed by the asymmetry of the error bars. From this experiment, we identified four key points that illustrate the superiority of the CWGP against GP. First, the proposed CWGP performed better than GP under all metrics considered (see Section \ref{sub:indices}). Second, the error bars and the noise variance are much tighter under CWGP, particularly around quarters number 50 and 200. Third, the proposed CWGP was able to successfully identify that the distribution of the signal cannot have negative support: even for ranges with missing data (see between quarters 150 and 200) the error bars did not reach zero. Fourth, CWGP was able to model positive deviations (see the peak around quarter 125) that fully contain the true process.

\begin{table}[ht]
	{\small
		\caption{Macroeconomic data: Performance of GP and CWGP.}
		\label{tab:table_macroeconomic} 
	}
	\centering
	\footnotesize
	\vspace{1em}
	\begin{tabular}{lrrrr}

		\toprule
		{} &         MAE &           MSE &      NLPD &         NLL \\
		\midrule
		GP  &        0.95 &        1.69 &        1.74 &       64.96 \\
		CWGP  &        \textbf{0.88} &        \textbf{1.75} &        \textbf{1.42} &       \textbf{57.36} \\
		\bottomrule
	\end{tabular}
\end{table}


\subsection{The Abalone, Ailerons and Creep datasets} 
\label{sub:real-world}

In this experiment we considered the three datasets originally used by WGP in \cite{warped04} and then by BWGP in \cite{bayesianwarped12} to validate CWGP. We considered the original WGP model with up to 3 non-linear components, and the proposed CWGP model with a maximum of 2 nonlinear components only. Notice that this follows the idea of compositional kernel search presented in \cite{DuvLloGroetal13}.

\subsubsection{Datasets and considered models} 
\label{sub:datasets_and_models_considered}

The regression problem associated to the Abalone dataset is to predict the age of an abalone (a type of sea snail) from 8-dimensional physical features. The Ailerons dataset is a simulated control problem where the aim is to predict the control action on the ailerons of an F16 aircraft from a 40-dimensional feature. In the Creep dataset the objective is to predict creep rupture stress (in MPa) for steel given the chemical composition and other 30-dimensional features. Following \cite{warped04,bayesianwarped12}, the training set sizes were chosen to be 1000 out of 4177, 1000 out of 7154 and 800 out of 2066 for Abalone, Ailerons and Creep datasets respectively. 

The models implemented were: (i) a standard GP, (ii) three variants of warped GP with one, two and three $\tanh(\cdot)$ components, and (iii) ten variants of the CWGP constructed by composing the elementary transformations presented in Section \ref{sec:transformations}. In total, 14 models were trained and evaluated, all of these used automatic relevance determination squared-exponential kernels \cite{Neal:ARD} and a constant mean function for the base (latent) Gaussian process. The motivation to implement ten variants of CWGP was to show the robustness of the proposed model to the choice of warpings in terms of both predictive performance and computational efficiency. All the experiments were implemented in {Python} using \textbf{G3py} \cite{g3py}, an open-source library for stochastic process modelling.

\setlength{\tabcolsep}{4.4pt}     
\renewcommand{\arraystretch}{0.65} 

\begin{table}[ht]
	{\small
		\caption{Performance of non-Gaussian models for the Abalone dataset: Training time (TimeT), evaluation time (TimeE), RMSE, MAE and NLPD. The first model is a GP; WP1, WGP2 and WGP3 are WGP models with one, two and three components respectively; and the remaining models are different variants of the proposed CWGP composed by the following elementary transformations: SA:SinhArcsinh, BC:Box-Cox, A:Arcsinh, L:affine, S:shifted. Times are measured in seconds and recall that the lower the score the better the model.}
		\label{tab:table_results_aba} 
	}
	\centering
	\footnotesize
	\begin{tabularx}{0.45\textwidth}{p{6em}|rr|rrr}
		\multicolumn{6}{c}{} \\ 
		\toprule
		{\textbf{Abalone}} &  TimeT &  TimeE &   RMSE &    MAE &   NLPD  \\
		\midrule
		GP                          &     19.927 &     1.362 &\textbf{2.158}&  1.543 &  2.287 \\
		WGP1                        &    103.55 &    82.94 &  2.164 &  1.534 &  2.189 \\
		WGP2                        &    124.57 &    72.48 &  2.174 &  1.526 &  2.079 \\
		WGP3                        &    127.93 &    84.98 &  2.181 &  1.539 &  2.200 \\
		SA 		                 	&     15.112 &     1.374 &  2.191 &  1.516 &  2.190 \\
		BC-L                		&     17.226 &     1.383 &  2.201 &  1.525 &  2.181 \\
		A-L               			&     23.552 &     1.385 &  2.223 &\textbf{1.512}&  2.073 \\
		BC-S               			&\textbf{10.811}&  1.382 &  2.211 &  1.530 &  2.225 \\
		BC-L-SA						&     16.456 &     1.395 &  2.465 &  1.561 &  5.272 \\
		BC-S-SA						&     11.354 &     1.380 &  3.980 &  1.525 &  2.295 \\
		A-L-BC-L					&     23.101 &     1.373 &  2.576 &  1.514 &\textbf{2.042}\\
		BC-L-A-L					&     16.236 &     1.396 &  2.295 &  1.547 &  2.263 \\
		A-L-BC-S					&     24.731 &\textbf{1.361}&2.302&  1.516 &  2.076\\
		BC-S-A-L					&     19.215 &     1.375 &  2.490 &  1.517 &  2.115\\
		\bottomrule
	\end{tabularx}
\end{table}
\begin{table}[ht]
	{\small
		\caption{Performance of non-Gaussian models for the Ailerons datasets. Notation follows that of Table \ref{tab:table_results_aba}.}
		\label{tab:table_results_ail} 
	}
	\centering
	\footnotesize
	\begin{tabularx}{0.45\textwidth}{p{6em}|rr|rrr}
	
		\multicolumn{6}{c}{} \\ 	
		\toprule
		{\textbf{Ailerons}} &  TimeT &  TimeE &   RMSE &    MAE &   NLPD  \\
		\midrule
		GP                          &     23.880 &     8.189 &  1.814 &  1.268 &  1.941\\
		WGP1                        &    151.571 &   239.947 &  1.800 &  1.264 &  1.935 \\
		WGP2                        &    160.557 &   229.789 &  1.739 &  1.231 &  1.881 \\
		WGP3                        &    179.417 &   245.485 &  1.765 &  1.247 &  1.903\\
		SA 		                 	&\textbf{11.523}&    10.274 &  1.876 &  1.258 &  1.821\\
		BC-L                		&     22.708 &     7.948 &  1.741 &  1.228 &  1.810\\
		A-L               			&     24.892 &     9.447 &  1.959 &  1.385 &  1.919\\
		BC-S               			&     20.001 &\textbf{6.992}&\textbf{1.702}&\textbf{1.210} &  1.815\\
		BC-L-SA						&     12.427 &    10.472 &  1.909 &  1.296 &  1.820 \\
		BC-S-SA						&     14.587 &     8.752 &  2.009 &  1.334 &  1.866\\
		A-L-BC-L					&     19.113 &     8.266 &  1.733 &  1.224 &  1.793\\
		BC-L-A-L					&     17.277 &     7.299 &  1.727 &  1.223 &\textbf{1.791}\\
		A-L-BC-S					&     18.417 &     7.023 &  1.707 &  1.212 &\textbf{1.791}\\
		BC-S-A-L					&     20.225 &     8.223 &  1.725 &  1.223 &  1.816\\
		\bottomrule
	\end{tabularx}
\end{table}
\begin{table}[ht]
	{\small
		\caption{Performance of non-Gaussian models for the Creep datasets. Notation follows that of Table \ref{tab:table_results_aba}.}
		\label{tab:table_results_creep} 
	}
	\centering
	\footnotesize
	\begin{tabularx}{0.45\textwidth}{p{6em}|rr|rrr}
		\multicolumn{6}{c}{} \\ 	
		\toprule
		{\textbf{Creep}} &  TimeT &  TimeE &   RMSE &    MAE &   NLPD  \\
		\midrule
		GP                          &     12.711 &     1.312 &  3.163 &  2.123 &  2.462 \\
		WGP1                        &     58.281 &    19.060 &\textbf{2.750}&  1.813 &  2.162\\
		WGP2                        &     73.323 &    29.419 &  2.758 &\textbf{1.808}&  2.166\\
		WGP3                        &     82.402 &    30.223 &  2.777 &  1.822 &  2.167\\
		SA 		                 	&     14.325 &     0.918 &  2.813 &  1.826 &\textbf{2.148}\\
		BC-L                		&      9.058 &     1.426 &  3.222 &  2.092 &  2.268\\
		A-L               			&     14.157 &     1.024 &  2.909 &  1.907 &  2.218\\
		BC-S               			&      8.139 &     1.582 &  3.076 &  2.055 &  2.325\\
		BC-L-SA						&     10.401 &     0.828 &  3.592 &  2.374 &  2.378 \\
		BC-S-SA						&\textbf{6.759}&     1.269 &  3.879 &  2.416 &  2.434\\
		A-L-BC-L					&     12.845 &\textbf{0.912}&  3.281 &  2.088 &  2.269\\
		BC-L-A-L					&     11.161 &     1.002 &  3.252 &  2.103 &  2.296 \\
		A-L-BC-S					&     11.191 &     1.503 &  3.207 &  2.026 &  2.236 \\
		BC-S-A-L					&      8.117 &     0.917 &  4.231 &  2.447 &  2.399 \\
		\bottomrule
	\end{tabularx}
\end{table}

\begin{figure*}[t]
	\centering
	\includegraphics[height=0.35\textwidth, trim=0 0 0 0, clip]{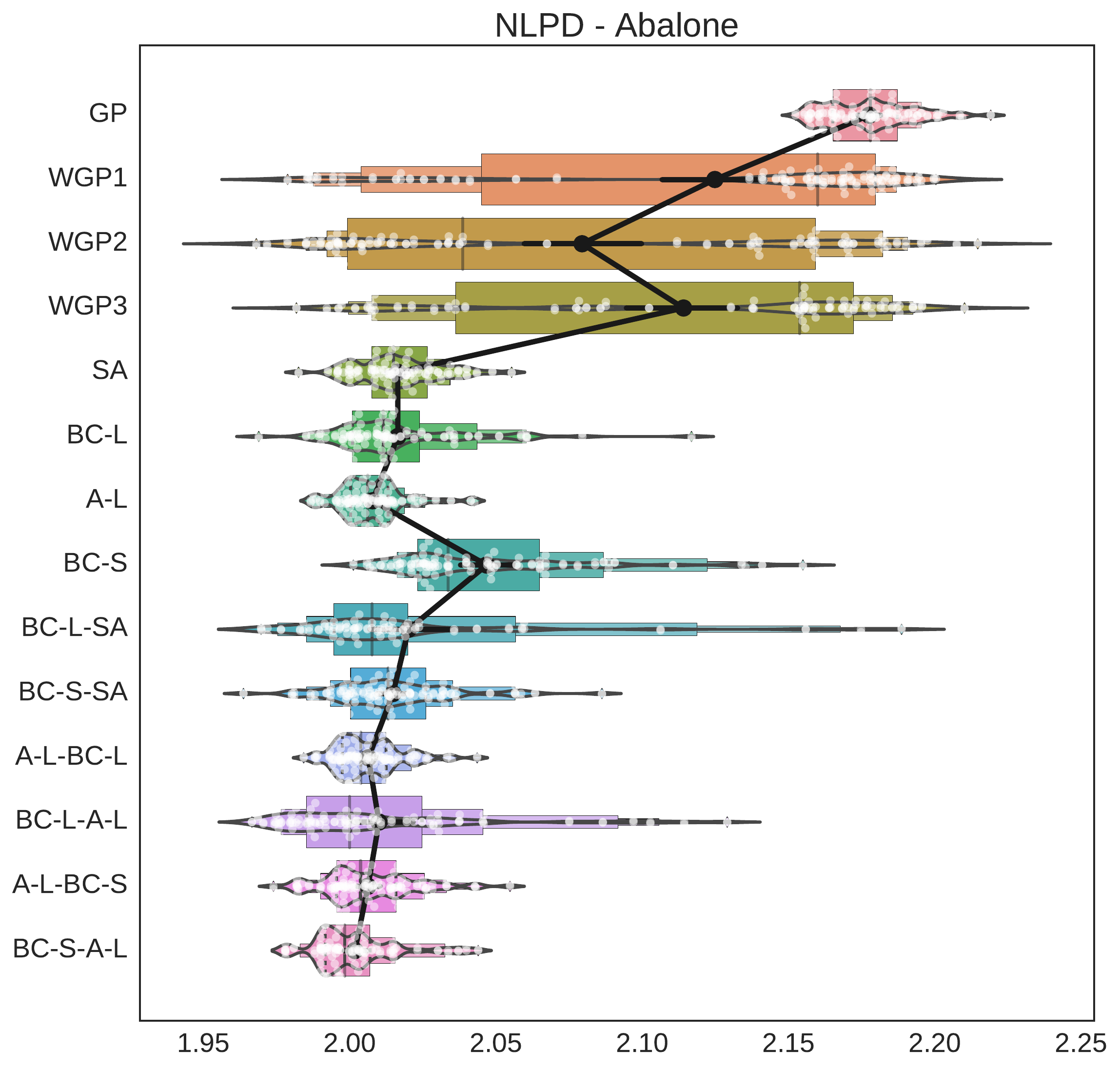}
	\hspace{-0.8em}
	\includegraphics[height=0.35\textwidth, trim=0 0 0 0, clip]{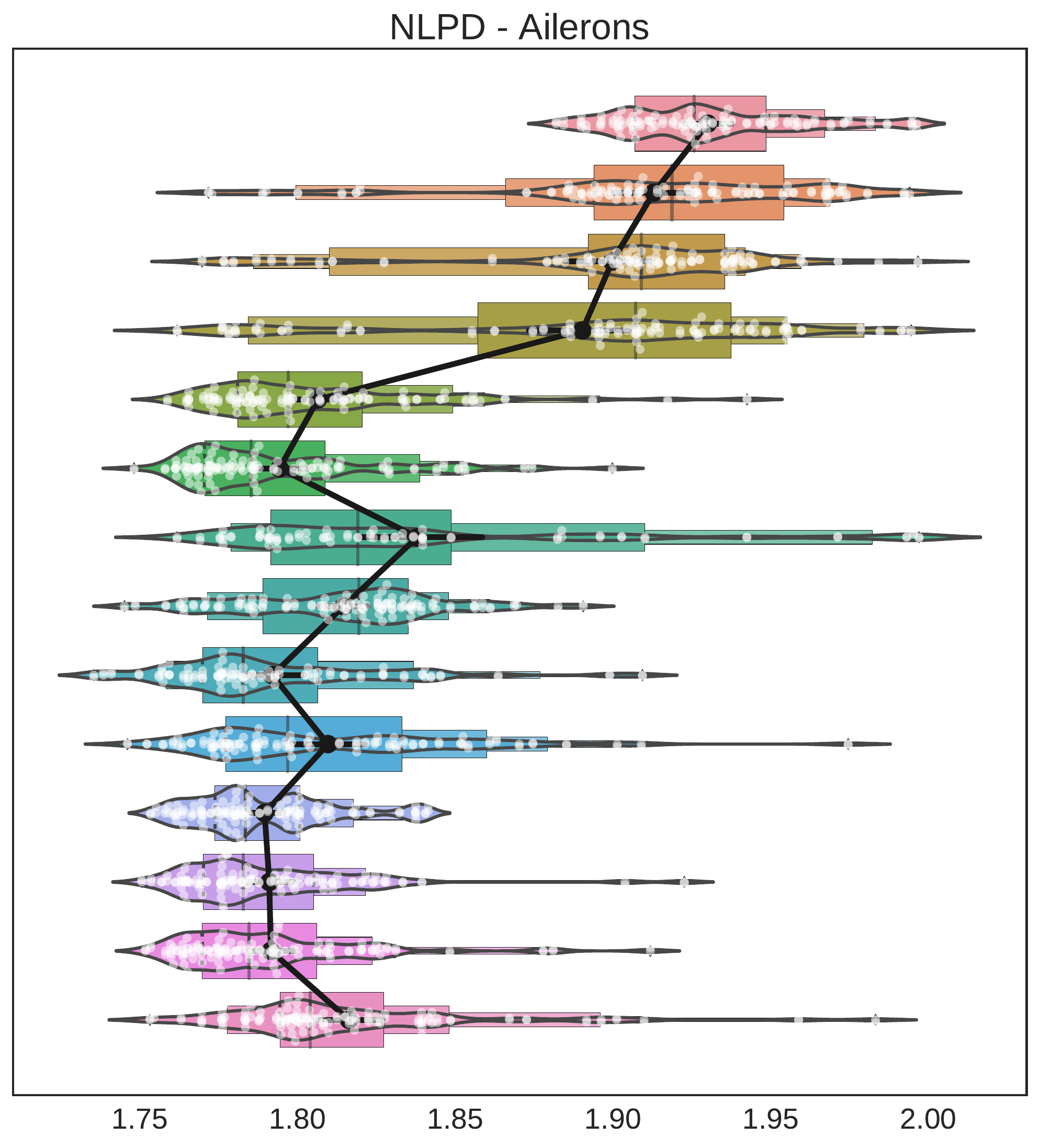}
	\hspace{-0.6em}
	\includegraphics[height=0.35\textwidth, trim=0 0 5 0, clip]{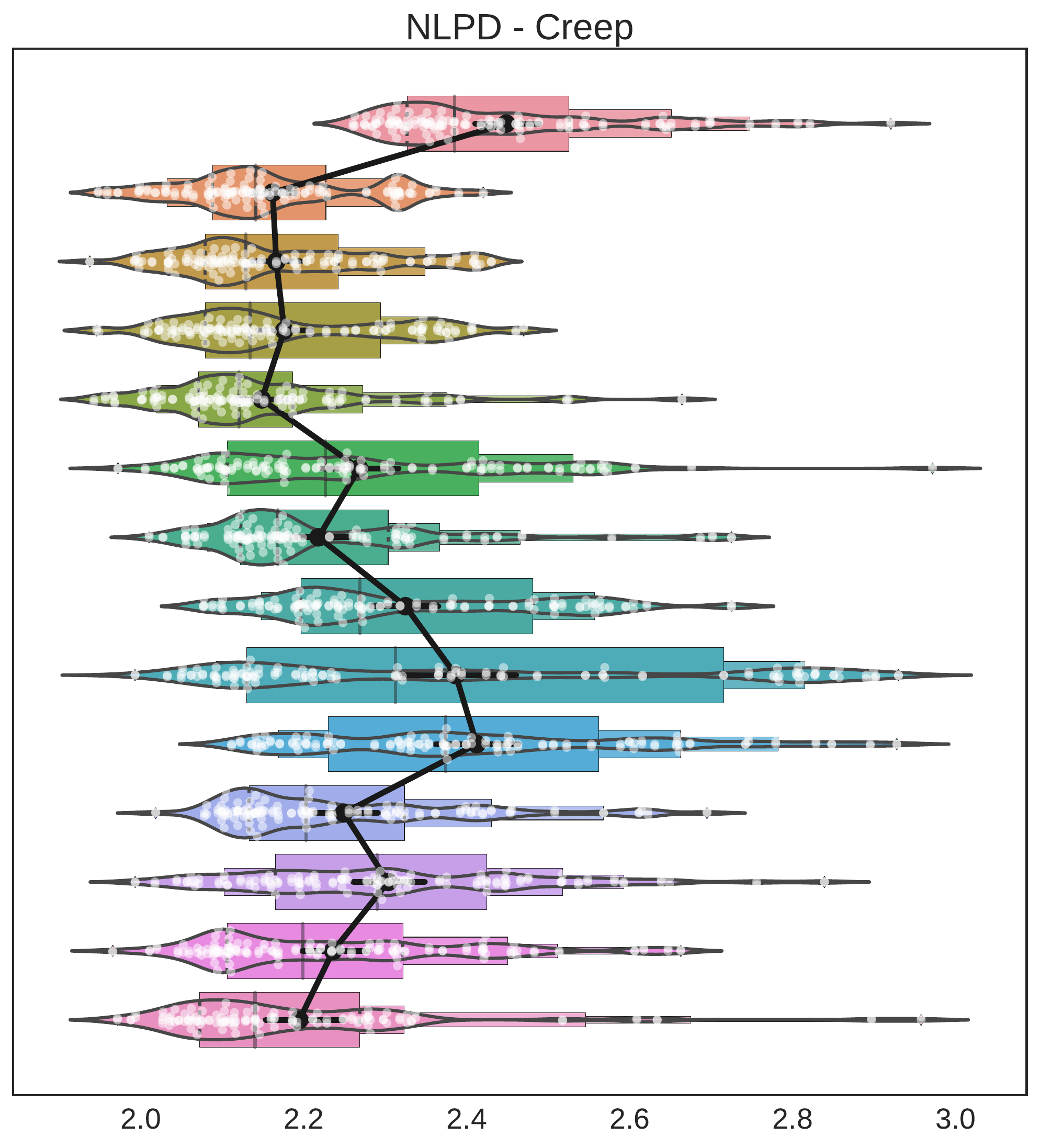}
	\caption{NLPD histograms (65 runs) for all models considered and the Abalone, Ailerons and Creep datasets. The white points are the scores, the black points are the average scores per model, and the boxes denote the quantiles. The models with more white points to the left-hand side of the plot are the the better ones.}
	\label{fig:NLPD_Abalone}
\end{figure*}

\subsubsection{Learning the latent GPs and the transformations} 
\label{sub:learning_the_latent_gp_hyperparameters_and_the_transformations_parameters}

For each model training was as follows. We randomly split the training set in two: An evaluation set and a validation set, both of same size. We minimised the NLL in eq.~\eqref{eq:NLL} with respect to the evaluation set using the BFGS method starting from 6 initial values of the (hyper)parameters: A default value independent of observations, a value calculated from the observations, a \textit{prelearning} value calculated using the trained standard GP, and three random values. We then selected the best model among the 6 trained models according to their RMSE in eq.~\eqref{eq:RMSE} over the validation set. This procedure was repeated 65 times for each  model and dataset in order to obtain an empirical distribution of the performance indices for each considered model.

\subsubsection{Model evaluation} 
\label{sub:model_evaluation}

We evaluated each selected model using the NLL, RMSE, MAE and NLPD indices in Section \ref{sub:indices} over the evaluation set. Tables \ref{tab:table_results_aba}-\ref{tab:table_results_creep} show the training and evaluation average times (TimeT and TimeE respectively) and the average values of all the performance indices considered for the 65 runs for all models and datasets. The proposed CWGP outperformed all models according to the NLPD, a non-Gaussian performance indicator, whereas GP and WGP performed better than CWGP in three cases according to RMSE/MAE. We attribute this to the Gaussian nature of RMSE/MAE that neglects asymmetry or kurtosis. 

Observe the appealing training and evaluation times of CWGP. In fact, notice that CWGP's training time was in the same order as that of the standard GP and sometimes even lower, this is because fitting a Gaussian model to non-Gaussian data might yield a flat NLL and therefore minimisation require several steps of BFGS. Fig.~\ref{fig:NLPD_Abalone} shows a histogram of the 65 NLPD scores for each model and dataset, where the white points are the scores, the black points are the average score per model and the boxes denote the quantiles. All non-Gaussian models outperform the standard GP in average, and we can see that WGP scores  have two modes (especially in the Abalone and Aileron datasets): one closer to the standard GP and another one closer to the scores of the proposed CWGP. This is due to the difficulty of training WGP, where in a number of cases the combination of the Newton-Raphson approximation and the BFGS optimiser fails to find the correct nonlinear map, therefore, the sum-of-tanh warping reduces to the identity and thus WGP collapses to the standard GP.


\section{Discussion and further work}
\label{sec:discussion}

We have provided a theoretically-grounded presentation of non-Gaussian processes resulting from nonlinear transformations of GPs using the change of variables theorem, thus complementing existing approaches such as WGP \cite{warped04}, Bayesian WGP \cite{bayesianwarped12} and deep GP \cite{deep_GP}. Although the warping functions considered by the aforementioned models can be chosen to be arbitrarily complex, their inverse and derivative require expensive numerical approximations. This motivated us to propose the compositionally-warped GP (CWGP), a variant of WGP that uses transformations given by compositions of multiple analytically-invertible and differentiable functions. Due to the expressiveness of the deep composition of elementary functions, the proposed CWGP model represents an improvement in terms of modelling ability with minimal numerical approximations, thus being a competitive alternative to existing methods. Furthermore, the proposed framework has been equipped with a set of explicit elementary functions, for which we have discussed their statistical properties (e.g., skewness and kurtosis), advantages in numerical computations and the expressiveness that can be achieved when using them as layers in a deep compositional warping. 

CWGP can be understood as a single, infinitely-wide, layer (the GP) fed into a deep composition of invertible single-neuron layers (the warping) that transforms non-Gaussian observations into Gaussian ones---see Fig.~\ref{fig:wgp}. This architecture is flexible in the sense that it admits arbitrary invertible and differentiable neurons as building blocks, where we have also shown that the model can operate as a \emph{black-box} model and the user only needs to choose the number of layers rather than the specific transformations (Sec.~\ref{sub:replicating_wgp}). Training the proposed CWGP via maximum likelihood is equivalent to optimising the network with respect to both the warping parameters and the probability of the transformed data---which is considered to be Gaussian. While coordinate-wise transformations can modify the marginal distributions, more general transformations allow us to go beyond and modify the dependency structure for the process, also known as copulas \cite{wilson2010copula}. 

Using the synthetic and real-world datasets in \cite{warped04,bayesianwarped12,tb3ms}, as well as financial and astronomical time series,  we have validated the proposed model against GP and WGP. We have shown that CWGP (i) succeeded in identifying key non-Gaussian statistical properties, (ii) outperformed both GP and WGP in terms of the distribution prediction error (NLPD), (iii) had lower computational complexity than WGP in experimental and theoretical terms and (iv) was robust to overfitting while becoming sparse, in terms of the number of active components,  when needed. Our findings motivate further research into (i) an automatic search for the optimal number and ordering of the compositional transformations as in \cite{DuvLloGroetal13}, (ii) constructing elementary transformations beyond the coordinate-wise case, e.g., via convolutions \cite{gpmm17,alvarez_lawrence,nips15}, thus making the warping structure both deep and wide; and finally (iii) depart from the maximum-likelihood metric to train non-Gaussian models and consider natural distances among distributions as those studied in Optimal Transport theory \cite{villani2008optimal,BLWB,Marzouk2016}.

\section*{Acknowledgements}
This work was funded by projects Conicyt-PCHA Doctorado Nacional 2016-21161789, Fondecyt-Iniciación 11171165 and Conicyt-PIA AFB170001 CMM.

\bibliography{library}

\vfill

\end{document}